\documentclass{article}

\PassOptionsToPackage{numbers,compress}{natbib}
\usepackage[preprint]{neurips_2026}
\makeatletter
\renewcommand{\@noticestring}{Preprint. Accepted at NeurIPS 2026.}
\makeatother

\usepackage[utf8]{inputenc} 
\usepackage[T1]{fontenc}    
\usepackage{hyperref}       
\usepackage{url}            
\usepackage{booktabs}       
\usepackage{amsfonts}       
\usepackage{nicefrac}       
\usepackage{microtype}      
\usepackage{xcolor}         

\usepackage{enumitem}
\usepackage{algorithm}
\usepackage{algpseudocode}
\usepackage{bm}
\usepackage{amsmath}
\usepackage{amssymb}
\usepackage{graphicx}

\usepackage[table]{xcolor}

\definecolor{mccbg}{RGB}{217,232,252}
\definecolor{elsbg}{RGB}{219,242,220}
\definecolor{ssrtbg}{RGB}{252,228,205}

\title{Revisiting Diffusion Fine-Tuning for Unsupervised Domain Adaptation}

\author{
Xuan Qi\textsuperscript{1,2} \quad
Yi Wei\textsuperscript{3,4}\thanks{Corresponding author.} \quad
Daniele Berardini\textsuperscript{1} \quad
Vito Paolo Pastore\textsuperscript{1,5} \quad
Vittorio Murino\textsuperscript{1,6}
\\[6pt]
\textsuperscript{1}AI for Good (AIGO), Istituto Italiano di Tecnologia, Genoa, Italy
\\[1pt]
\textsuperscript{2}DITEN, University of Genoa, Genoa, Italy
\\[1pt]
\textsuperscript{3}National Key Laboratory of Novel Software Technology, Nanjing University, China
\\[1pt]
\textsuperscript{4}School of Intelligence Science and Technology, Nanjing University, China
\\[1pt]
\textsuperscript{5}MaLGa, DIBRIS, University of Genoa, Genoa, Italy
\\[1pt]
\textsuperscript{6}Department of Computer Science, University of Verona, Verona, Italy
\\[5pt]
\texttt{\{xuan.qi, daniele.berardini, vittorio.murino\}@iit.it}
\\[2pt]
\texttt{ywei@smail.nju.edu.cn}
\qquad
\texttt{vito.paolo.pastore@unige.it}
}

\begin{document}

\maketitle

\begin{abstract}
Diffusion-based unsupervised domain adaptation (UDA) improves cross-domain transfer by generating target-specific synthetic data for downstream adaptation. Existing methods are largely designed for single-target adaptation: when a model trained on one labeled source domain must be adapted to multiple unlabeled target domains, they typically require separate diffusion fine-tuning for each source--target pair, causing training, storage, and deployment costs to grow with the number of targets. In this paper, we study multi-target data generation for diffusion-based UDA, where a single source-guided diffusion fine-tuning process is reused to generate target-specific synthetic data for multiple target domains. We propose \textsc{MUSE} (\textbf{M}ulti-target \textbf{U}DA-oriented \textbf{S}ynthesis with \textbf{E}fficient diffusion fine-tuning), a decoupled adaptation framework that separates source-supervised semantic adaptation from target-specific style adaptation. MUSE uses a shared semantic branch updated by labeled source data and target-private style branches specialized to individual target domains, enabling target-specific generation while avoiding repeated source-guided fine-tuning for each target. Experiments on standard UDA benchmarks show that MUSE achieves a stronger accuracy--efficiency trade-off than repeated per-target diffusion adaptation, reducing diffusion fine-tuning cost while improving average target-domain accuracy. The project page is available at \url{https://xuanqi99.github.io/MUSE/}.
\end{abstract}

\section{Introduction}

Unsupervised domain adaptation (UDA) aims at adapting a model trained on a labeled source domain to an unlabeled target domain under distribution shift. Most classical UDA methods address this problem in feature space by learning transferable or domain-invariant representations, for example through discrepancy minimization~\cite{pan2011domain,tzeng2014deep,long2015learning,sun2016deepcoral}, adversarial alignment~\cite{ganin2015unsupervised,ganin2016domain,pei2018mada,chen2019batch,zhang2023els}, conditional alignment~\cite{long2017jan,long2018conditional,jiang2020implicit}, reconstruction and image-level translation~\cite{long2016residual,bousmalis2017pixelda,hoffman2018cycada}, pseudo-labeling and self-training~\cite{saito2017asymmetric,kumar2020gradual}, or margin- and decision-boundary-aware regularization~\cite{saito2018maximum,xie2018semantic,jin2020minimum}. More recently, diffusion models have opened a complementary data-centric route to UDA. Instead of adapting only the discriminative model, one can reduce domain shift at the sample level by generating target-relevant data, for example cross-domain diffusion adaptation~\cite{zhuang2024terra}, or class-conditional target-domain synthesis for downstream adaptation~\cite{zhang2025dcdm}.

This generative perspective has become increasingly practical for two reasons. First, modern diffusion models provide strong image priors and flexible conditioning mechanisms~\cite{ho2020denoising,song2021score,nichol2021improved,dhariwal2021diffusion,rombach2022high,nichol2022glide}, making it possible to generate samples that preserve task-relevant semantics while moving toward target-domain appearance. Second, parameter-efficient adaptation methods such as adapters (e.g., LoRA~\cite{hu2022lora}) enable large diffusion backbones to be customized with modest additional training and storage cost~\cite{houlsby2019parameter,hu2022lora,karimimahabadi2021compacter}. Together, these advances make diffusion-based synthetic-data generation a practical data-centric complement to discriminative UDA, providing target-relevant synthetic supervision for downstream adaptation.

However, most existing data-centric transfer methods are still designed for a \emph{single-target} regime. Actually, many of such earlier 
approaches typically constructed a bridge between one source and one target through cross-domain translation or intermediate-domain generation~\cite{zhu2017cyclegan,hoffman2018cycada,gong2019dlow}. Recent diffusion-based UDA methods largely inherit the same structure: they adapt a generator for a specific source--target pair, and then use the adapted generator only for that target~\cite{zhuang2024terra,zhang2025dcdm}. This workflow is natural when only one target domain is of interest. In many practical settings, however, one labeled source domain must support adaptation to multiple unlabeled target domains. In such cases, 
adapting a separate diffusion model for each target causes training cost and deployment complexity to grow with the number of targets. More importantly, it repeatedly relearns source-side class semantics that are shared across targets, while target-specific adaptation is still needed to capture domain-specific variations.

Motivated by this limitation, we study \emph{multi-target data generation for diffusion-based UDA}, where a single labeled source domain is used to support target-specific synthetic-data generation for multiple unlabeled target domains sharing the same label space. 
Rather than training a separate diffusion model for each source--target pair, we focus on the generator adaptation stage and ask whether diffusion adaptation can be decomposed into reusable source-supervised semantics and target-specific appearance specialization. 
Such a decomposition would allow one source-guided fine-tuning process to serve multiple targets while preserving target specificity and reducing cross-target interference.

In this work, we propose \textsc{MUSE} (\textbf{M}ulti-target \textbf{U}DA-oriented \textbf{S}ynthesis with \textbf{E}fficient diffusion fine-tuning), a parameter-efficient diffusion adaptation framework for multi-target data generation in UDA. 
MUSE is built on a shared-private decomposition that separates source-supervised semantic adaptation from target-specific style adaptation. 
Specifically, a shared semantic branch learns class-discriminative information from labeled source data, while lightweight target-private style branches capture the appearance of individual target domains. 
Together with branch-decoupled optimization, this design enables one source-guided diffusion fine-tuning process to generate target-specific bridge data for multiple targets, while mitigating cross-target interference.

\begin{itemize}
    \item We introduce \emph{multi-target data generation} for diffusion-based UDA, where a single labeled source domain supports target-specific synthetic data generation for multiple unlabeled target domains. This setting avoids performing separate source-involved diffusion fine-tuning for each source--target pair, thereby reducing the training and storage costs of diffusion-based adaptation.

    \item We propose \textsc{MUSE}, a novel parameter-efficient multi-target diffusion adaptation 
    framework that combines a source-shared branch, target-specific private branches, and 
    decoupled optimization for joint adaptation across targets.

    \item We show that \textsc{MUSE} achieves a better accuracy--efficiency trade-off than separate per-target diffusion fine-tuning for downstream UDA. In particular, it reduces diffusion fine-tuning time and the number of additional trainable adapter parameters while improving average target-domain accuracy.
\end{itemize}

\section{Related Work}

\paragraph{Unsupervised domain adaptation.}
Prior UDA research has been dominated by discriminative adaptation in feature space. Representative directions include discrepancy-based alignment~\cite{pan2011domain,tzeng2014deep,long2015learning,sun2016deepcoral}, adversarial alignment~\cite{ganin2015unsupervised,ganin2016domain,pei2018mada,chen2019batch,zhang2023els}, conditional or class-aware alignment~\cite{long2017jan,long2018conditional,jiang2020implicit}, reconstruction and image-level adaptation~\cite{long2016residual,bousmalis2017pixelda,hoffman2018cycada}, pseudo-labeling and self-training~\cite{saito2017asymmetric,kumar2020gradual}, and margin- or decision-boundary-aware regularization~\cite{saito2018maximum,xie2018semantic,jin2020minimum}. Beyond the standard single-source setting, multi-source and decentralized variants have also been studied to transfer knowledge across multiple domains~\cite{zhao2018adversarial,peng2019moment,feng2021kd3a}. More recently, multi-target UDA has considered adapting a single source-trained model to multiple unlabeled target domains~\cite{shin2025merge,li2024training,monga2024cosmo}. These methods primarily improve transfer by learning more invariant or transferable representations for a fixed discriminative model. In contrast, our work takes a data-centric view and studies how to adapt a generator to produce target-specific synthetic supervision for downstream UDA.

\paragraph{Diffusion fine-tuning.}
Diffusion models have become a strong foundation for conditional image generation~\cite{ho2020denoising,song2021score,nichol2021improved,dhariwal2021diffusion,rombach2022high,nichol2022glide}. In practice, adapting such models often relies on parameter-efficient fine-tuning (PEFT) rather than full fine-tuning. In the broader PEFT literature, adapters and low-rank updates provide lightweight alternatives to full parameter updates~\cite{houlsby2019parameter,hu2022lora,karimimahabadi2021compacter}, and recent variants further expand this design space through higher-rank or more structured parameterizations, such as HiRA~\cite{huang2025hira} and QR-LoRA~\cite{yang2025qrlora}. In diffusion models, PEFT-style adaptation has been used for personalization, editing, controllable generation, and modular customization, including Textual Inversion~\cite{gal2023textualinversion}, DreamBooth~\cite{ruiz2023dreambooth}, Custom Diffusion~\cite{kumari2023customdiffusion}, Prompt-to-Prompt~\cite{hertz2023prompttoprompt}, Null-Text Inversion~\cite{mokady2023nulltext}, InstructPix2Pix~\cite{brooks2023instructpix2pix}, ControlNet~\cite{zhang2023controlnet}, SVDiff~\cite{han2023svdiff}, BLIP-Diffusion~\cite{li2023blipdiffusion}, T2I-Adapter~\cite{mou2024t2iadapter}, and Orthogonal Adaptation~\cite{po2024orthogonal}. Despite sharing parameter-efficient diffusion adaptation, MUSE targets a different setting: instead of customizing a generator for a single concept or a single target domain, we study how one source-guided adaptation process can be reused across multiple target domains. Accordingly, we separate reusable source-side semantics from target-specific components within a single fine-tuning framework.

\paragraph{Diffusion-based UDA.}
Data-centric domain transfer predates diffusion models: image translation and intermediate-domain generation have long been used to mitigate domain shift via synthetic bridge samples~\cite{zhu2017cyclegan,hoffman2018cycada,gong2019dlow,qi2026vtduda}. Recent diffusion-based methods revisit this direction with stronger generative priors and parameter-efficient adaptation. DCDM studies class-conditional target-style synthesis for downstream adaptation~\cite{zhang2025dcdm}; and Terra develops time-varying low-rank adaptation for cross-domain diffusion fine-tuning, with applications to target-sample generation and source-to-target transformation~\cite{zhuang2024terra}. In particular, Terra models diffusion adaptation for each source--target pair through time-varying low-rank updates, whereas MUSE shares source-supervised semantic adaptation across multiple targets while retaining target-specific style branches within a single fine-tuning process. By contrast, we study multi-target data generation for diffusion-based UDA, where one labeled source domain supports target-specific synthetic bridge-data generation for multiple unlabeled target domains without adapting a separate generator for each source--target pair.

\section{Method}
\label{sec:method}

\subsection{Problem Setup}
\label{sec:problem_setup}

We study \emph{multi-target data generation} for diffusion-based UDA. The input consists of one labeled source domain \(\mathcal{D}_s=\{(x_i^s,y_i^s)\}_{i=1}^{N_s}\) and \(M\) unlabeled target domains \(\{\mathcal{D}_t^{(m)}\}_{m=1}^M\), where \(\mathcal{D}_t^{(m)}=\{x_j^{t,m}\}_{j=1}^{N_t^{(m)}}\). We assume the standard closed-set setting: all domains share the same label space \(\mathcal{Y}\), while their image distributions differ.

Our focus is the \emph{generator adaptation stage}. Let \(D_{\theta_0}\) denote the denoising prediction network of a pre-trained latent diffusion model with frozen backbone parameters \(\theta_0\), and let \(\phi\) denote the trainable adaptation parameters introduced during fine-tuning. Following latent diffusion, we work in the latent space of a frozen VAE encoder \(E\). For an input image \(x\), we write \(z_0=E(x)\), and construct a noisy latent at timestep \(t\in\{1,\dots,T\}\) by
\begin{equation}
z_t=\sqrt{\bar{\alpha}_t}\,z_0+\sqrt{1-\bar{\alpha}_t}\,\epsilon,
\qquad
\epsilon\sim\mathcal{N}(0,I),
\label{eq:forward_process}
\end{equation}
where \(T\) is the number of diffusion timesteps, \(\bar{\alpha}_t\) is the cumulative noise-schedule coefficient, and \(I\) is the identity covariance of matching latent dimension. We use \(D_{\theta_0,\phi}\) to denote the adapted denoising prediction network obtained by combining the frozen backbone \(\theta_0\) with the trainable parameters \(\phi\). Let \(\eta(z_0,t,\epsilon)\) denote the scheduler-dependent denoising target; in the standard noise-prediction case, \(\eta(z_0,t,\epsilon)=\epsilon\).

\begin{figure*}[t]
    \centering
    \includegraphics[width=0.98\textwidth]{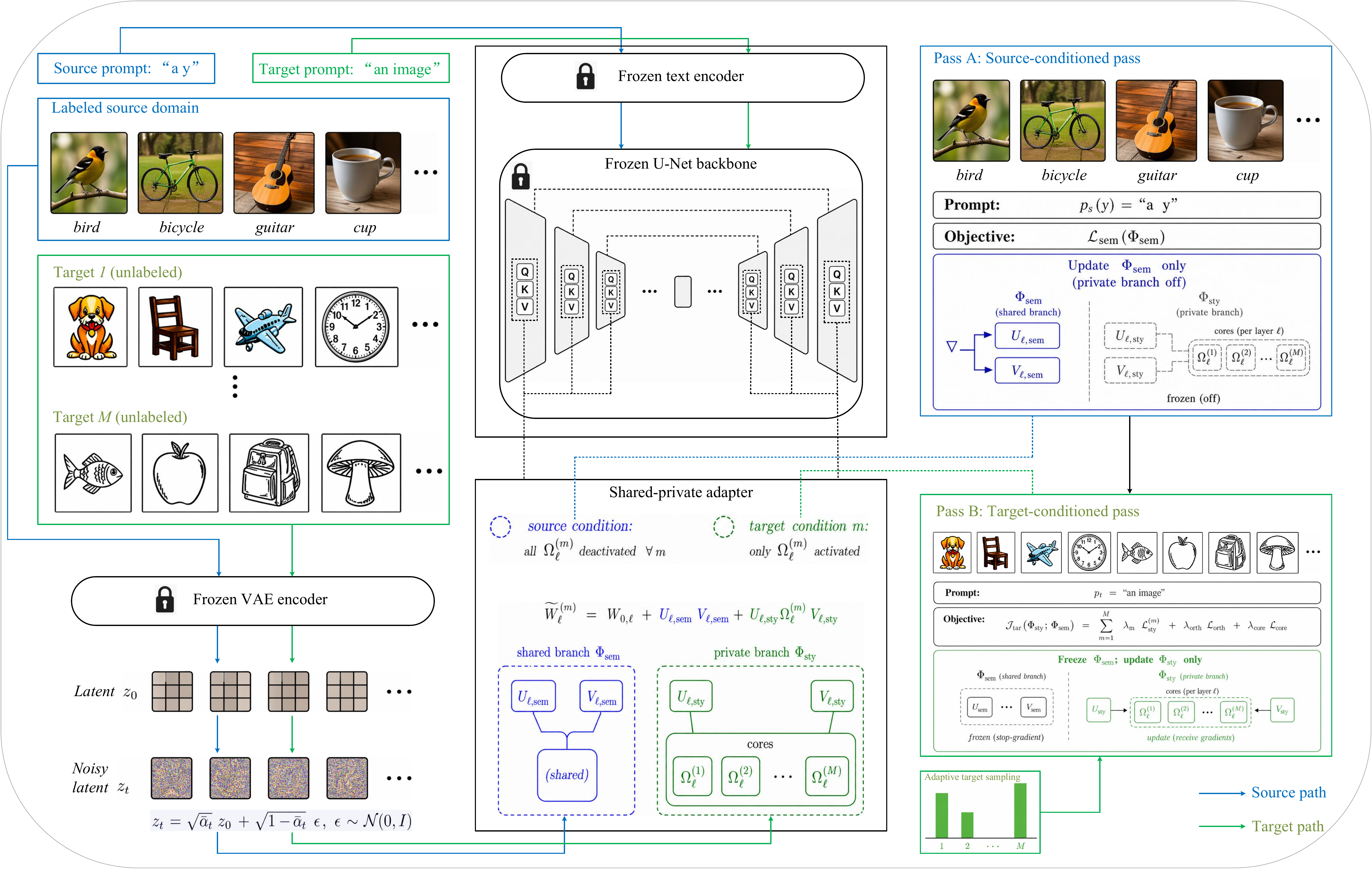}
    \caption{
    Overview of the proposed MUSE pipeline. MUSE performs branch-decoupled diffusion adaptation with a source-supervised shared semantic branch and target-private style branches, then activates the corresponding target branch to generate target-specific bridge samples for downstream UDA.
    }
    \label{fig:muse_pipeline}
\end{figure*}

After fine-tuning, the adapted generator is instantiated with a target index \(m\in\{1,\dots,M\}\) and used to produce a target-specific labeled bridge set
\(\widehat{\mathcal{B}}^{(m)}(\phi)=\{(\widehat{x}_i^{(m)},\widehat{y}_i^{(m)})\}_i\).
This bridge set contains labeled synthetic samples generated to reflect the visual characteristics of target domain \(m\), including source-to-target translated samples and class-conditioned samples defined in Sec.~\ref{sec:bridge_generation}. It is used as synthetic supervision for downstream adaptation on \(\mathcal{D}_t^{(m)}\). For each target domain \(m\), let \(f^{(m)}\) denote a downstream UDA learner trained using both \(\mathcal{D}_t^{(m)}\) and \(\widehat{\mathcal{B}}^{(m)}(\phi)\). We summarize the downstream role of the adapted generator as
\begin{equation}
f^{(m)\star}
\in
\arg\min_f
\mathcal{R}
\!\left(
f;
\mathcal{D}_t^{(m)},
\widehat{\mathcal{B}}^{(m)}(\phi)
\right),
\qquad m=1,\dots,M.
\label{eq:downstream}
\end{equation}
Here \(\mathcal{R}\) denotes the training objective of the downstream UDA method. Under this formulation, the goal of generator adaptation is to learn a single adapted generator that can be reused across target domains while producing target-specific labeled bridge data for downstream UDA. Figure~\ref{fig:muse_pipeline} gives an overview of the proposed MUSE pipeline.

\subsection{Shared-Private Adapter}

We instantiate the generator from a pre-trained SDXL~\cite{SDXL} latent diffusion model. The VAE encoder and text encoders are kept frozen, while trainable adaptation is restricted to the attention projections of the diffusion U-Net. Let \(\mathcal{L}\) denote the set of adapted attention projection layers. For each \(\ell\in\mathcal{L}\), let \(W_{0,\ell}\in\mathbb{R}^{d_\ell^{\mathrm{out}}\times d_\ell^{\mathrm{in}}}\) denote the frozen pre-trained weight. We parameterize the adapted weight for target domain \(m\) as
\begin{equation}
\widetilde{W}_\ell^{(m)}
=
W_{0,\ell}
+
U_{\ell,\mathrm{sem}}V_{\ell,\mathrm{sem}}
+
U_{\ell,\mathrm{sty}}\Omega_\ell^{(m)}V_{\ell,\mathrm{sty}},
\qquad m=1,\dots,M.
\label{eq:main_decomp}
\end{equation}
Here \(U_{\ell,\mathrm{sem}}\in\mathbb{R}^{d_\ell^{\mathrm{out}}\times r_{\mathrm{sem}}}\) and \(V_{\ell,\mathrm{sem}}\in\mathbb{R}^{r_{\mathrm{sem}}\times d_\ell^{\mathrm{in}}}\) define the shared semantic low-rank branch with rank \(r_{\mathrm{sem}}\). The target-conditioned term uses target-shared style projection factors \(U_{\ell,\mathrm{sty}}\in\mathbb{R}^{d_\ell^{\mathrm{out}}\times r_{\mathrm{sty}}}\) and \(V_{\ell,\mathrm{sty}}\in\mathbb{R}^{r_{\mathrm{sty}}\times d_\ell^{\mathrm{in}}}\), together with a target-private core \(\Omega_\ell^{(m)}\in\mathbb{R}^{r_{\mathrm{sty}}\times r_{\mathrm{sty}}}\), where \(r_{\mathrm{sty}}\) is the style rank.

For the source condition, all target-private cores are deactivated, so \(\widetilde{W}_\ell^{(s)}=W_{0,\ell}+U_{\ell,\mathrm{sem}}V_{\ell,\mathrm{sem}}\). For target domain \(m\), only the corresponding core \(\Omega_\ell^{(m)}\) is activated. We use a domain-conditioning index \(c\in\{s,1,\dots,M\}\), where \(c=s\) denotes the source condition and \(c=m\) denotes target domain \(m\). We denote the parameter collections by \(\Phi_{\mathrm{sem}}=\{U_{\ell,\mathrm{sem}},V_{\ell,\mathrm{sem}}\}_{\ell\in\mathcal{L}}\) and \(\Phi_{\mathrm{sty}}=\{U_{\ell,\mathrm{sty}},V_{\ell,\mathrm{sty}},\Omega_\ell^{(1)},\dots,\Omega_\ell^{(M)}\}_{\ell\in\mathcal{L}}\), so that \(\phi=(\Phi_{\mathrm{sem}},\Phi_{\mathrm{sty}})\). Relative to repeated source-involved fine-tuning for each source--target pair under this adapter form, MUSE stores the source-updated semantic branch and the style projection factors once, and introduces target-specific parameters through lightweight private cores. A formal comparison of trainable adapter parameter scaling is provided in Appendix~\ref{app:muse_scaling}.

\subsection{Training Objective and Two-Pass Optimization}

For source-domain inputs, we use class-conditional prompts \(p_s(y)=\texttt{``a }y\texttt{''}\). Under the source condition, all target-private cores are deactivated, so the source denoising loss depends only on \(\Phi_{\mathrm{sem}}\). For target-domain inputs, labels are unavailable, so we use a generic prompt \(p_t=\texttt{``an image''}\). Let \(\eta^s=\eta(z_0^s,t,\epsilon)\) and \(\eta^{t,m}=\eta(z_0^{t,m},t,\epsilon)\) denote the denoising targets for source and target samples. The source and target denoising losses are
\begin{align}
\mathcal{L}_{\mathrm{sem}}(\Phi_{\mathrm{sem}})
&=
\mathbb{E}_{(x^s,y^s)\sim\mathcal{D}_s,\;t,\;\epsilon}
\!\left[
w(t)
\left\|
\eta^s-D_{\theta_0,\phi}\!\left(z_t^s,t,p_s(y^s),s\right)
\right\|_2^2
\right],
\label{eq:lsem_main}
\\
\mathcal{L}_{\mathrm{sty}}^{(m)}(\Phi_{\mathrm{sem}},\Phi_{\mathrm{sty}})
&=
\mathbb{E}_{x^{t,m}\sim\mathcal{D}_t^{(m)},\;t,\;\epsilon}
\!\left[
w(t)
\left\|
\eta^{t,m}-D_{\theta_0,\phi}\!\left(z_t^{t,m},t,p_t,m\right)
\right\|_2^2
\right],
\label{eq:lsty_main}
\end{align}
where \(z_t^s\) and \(z_t^{t,m}\) are obtained by applying Eq.~\eqref{eq:forward_process} to source and target images, respectively; \(t\) is sampled from the diffusion timestep distribution; and \(w(t)\) is the timestep-dependent weighting.

To combine target-domain losses, we use nonnegative weights \(\lambda_m\ge 0\). We further regularize the target-private cores using an orthogonality regularizer and a magnitude regularizer:
\begin{align}
\mathcal{L}_{\mathrm{tar}}(\Phi_{\mathrm{sem}},\Phi_{\mathrm{sty}})
&=
\sum_{m=1}^{M}\lambda_m\,
\mathcal{L}_{\mathrm{sty}}^{(m)}(\Phi_{\mathrm{sem}},\Phi_{\mathrm{sty}}),
\label{eq:ltar_main}
\\
\mathcal{L}_{\mathrm{orth}}
&=
\sum_{\ell\in\mathcal{L}}\sum_{1\le i<j\le M}
\left\|(\Omega_\ell^{(i)})^\top\Omega_\ell^{(j)}\right\|_F^2,
\qquad
\mathcal{L}_{\mathrm{core}}
=
\sum_{\ell\in\mathcal{L}}\sum_{m=1}^{M}\|\Omega_\ell^{(m)}\|_F^2.
\label{eq:reg_terms_main}
\end{align}
Here \(\|\cdot\|_F\) denotes the Frobenius norm. 
The regularizers keep target-private cores diverse and bounded: \(\mathcal{L}_{\mathrm{orth}}\) penalizes aligned core coefficients across targets, while \(\mathcal{L}_{\mathrm{core}}\) limits their magnitude. 
Since the style projections are not orthonormal, \(\mathcal{L}_{\mathrm{orth}}\) promotes coefficient-space diversity rather than exact orthogonality of the induced full weight updates.

We optimize MUSE with branch-decoupled two-pass training. The source-side objective is
\(\mathcal{J}_{\mathrm{src}}(\Phi_{\mathrm{sem}})=\mathcal{L}_{\mathrm{sem}}(\Phi_{\mathrm{sem}})\). The target-side objective is
\begin{equation}
\mathcal{J}_{\mathrm{tar}}(\Phi_{\mathrm{sty}};\Phi_{\mathrm{sem}})
=
\mathcal{L}_{\mathrm{tar}}(\Phi_{\mathrm{sem}},\Phi_{\mathrm{sty}})
+
\lambda_{\mathrm{orth}}\mathcal{L}_{\mathrm{orth}}
+
\lambda_{\mathrm{core}}\mathcal{L}_{\mathrm{core}},
\label{eq:tar_obj_main}
\end{equation}
where \(\lambda_{\mathrm{orth}}\ge 0\) and \(\lambda_{\mathrm{core}}\ge 0\) control the strengths of the core-diversity and core-magnitude regularizers, respectively.

Each training iteration consists of one source-conditioned pass and one target-conditioned pass. In the source-conditioned pass, all target-private cores are deactivated, and gradients are routed only to the shared semantic branch \(\Phi_{\mathrm{sem}}\). In the target-conditioned pass, the semantic branch participates in the forward computation but is treated as stop-gradient, so gradients are routed only to the style parameters \(\Phi_{\mathrm{sty}}\). In practice, the target denoising loss is computed using a minibatch from one sampled target domain, while the core regularizers are evaluated over all target-private cores; details on the resulting effective target weighting are provided in Appendix~\ref{app:optimization_details}. This two-pass procedure decouples the optimization of source-supervised semantics and target-specific appearance: labeled source data updates the shared semantic branch, whereas unlabeled target data updates the target-conditioned style branch.

\subsection{Adaptive Target Sampling}

Target domains can differ in optimization difficulty during multi-target diffusion adaptation. To allocate more target-side updates to harder domains, we sample target domains adaptively according to their recent denoising losses. Let \(\bar{\ell}_m\) denote an exponential moving average of the target denoising loss for domain \(m\). When domain \(m\) is sampled, we update \(\bar{\ell}_m \leftarrow \beta \bar{\ell}_m + (1-\beta)\ell_m\), where \(\ell_m\) is the current minibatch estimate of the target denoising loss and \(\beta\in[0,1)\) is the smoothing coefficient. The moving averages of unsampled domains are kept unchanged. We then define the target-domain sampling distribution as
\begin{equation}
p_m
=
\frac{(\bar{\ell}_m+\tau)^\alpha}{\sum_{k=1}^{M}(\bar{\ell}_k+\tau)^\alpha},
\qquad m=1,\dots,M,
\label{eq:sampling_main}
\end{equation}
where \(\tau>0\) is a small constant for numerical stability and \(\alpha\ge 0\) controls the strength of loss-based prioritization. Under this scheme, target domains with persistently larger denoising losses are sampled more frequently during target-conditioned updates. Details on the effective target weighting induced by this sampling strategy are provided in Appendix~\ref{app:optimization_details}.

\subsection{Bridge Generation}
\label{sec:bridge_generation}

\begin{figure}[t]
    \centering
    \includegraphics[width=0.98\linewidth]{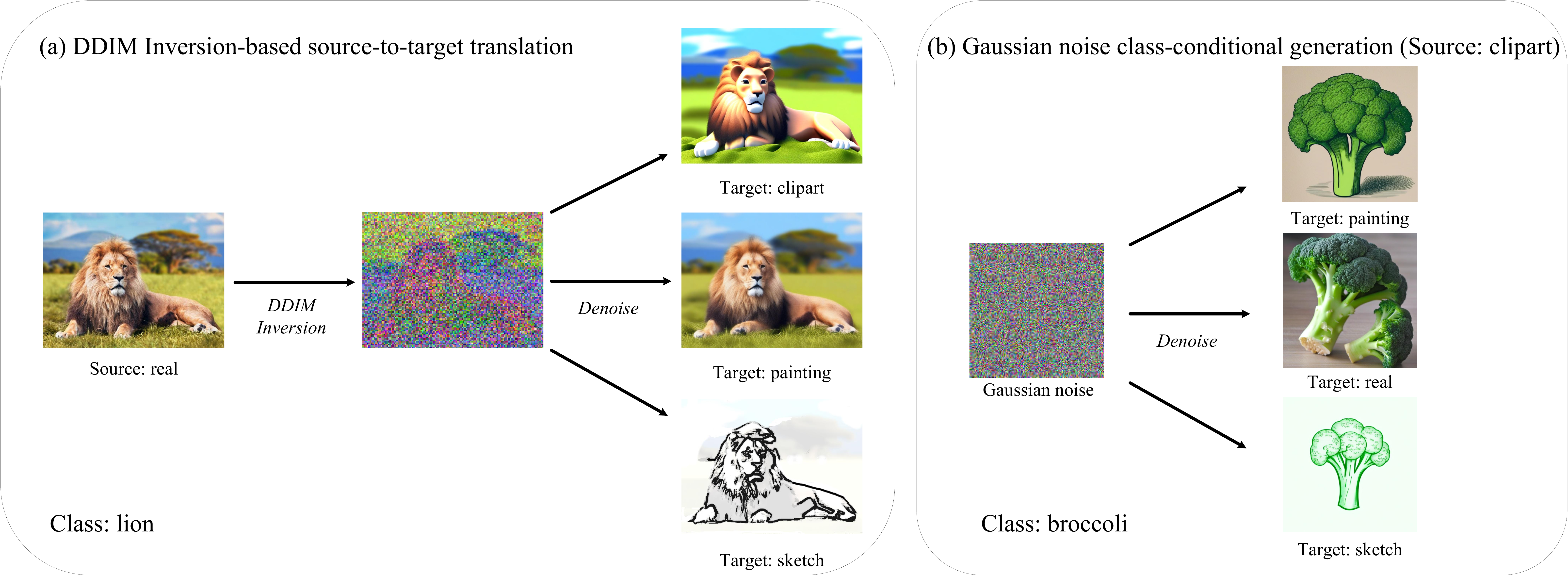}
    \caption{
    Bridge generation on miniDomainNet.
    (a) DDIM inversion translates a Real-domain \emph{lion} into Clipart, Painting, and Sketch targets, i.e., \(R\!\rightarrow\!\{C,P,S\}\).
    (b) Gaussian-noise class-conditional generation synthesizes \emph{broccoli} samples for Painting, Real, and Sketch targets under \(C\!\rightarrow\!\{P,R,S\}\).
    }
    \label{fig:bridge_generation_modes}
\end{figure}

After fine-tuning, the shared semantic branch and the style projection factors are reused for all target domains, and target domain \(m\) is selected by activating only its private cores \(\{\Omega_\ell^{(m)}\}_{\ell\in\mathcal{L}}\). For each target, MUSE constructs a labeled bridge set with two complementary components:
\begin{equation}
\widehat{\mathcal{B}}^{(m)}(\phi)
=
\widehat{\mathcal{B}}_{\mathrm{inv}}^{(m)}(\phi)
\cup
\widehat{\mathcal{B}}_{\mathrm{cls}}^{(m)}(\phi),
\label{eq:bridge_set_components}
\end{equation}
where the union denotes dataset concatenation. Here \(\widehat{\mathcal{B}}_{\mathrm{inv}}^{(m)}\) denotes labeled source-to-target translated samples obtained by DDIM inversion: a labeled source image \((x^s,y^s)\) is inverted and then denoised with the target-specific branch for domain \(m\), yielding a target-style sample assigned the source label \(y^s\). Complementarily, \(\widehat{\mathcal{B}}_{\mathrm{cls}}^{(m)}\) denotes labeled class-conditional target-style samples generated directly from Gaussian noise using the target-specific branch for domain \(m\) and class prompts \(p_s(y)=\texttt{``a }y\texttt{''}\). Each generated sample is assigned the class label specified by its prompt. The former provides instance-preserving target-style translations, while the latter increases class-conditioned target-style diversity. Figure~\ref{fig:bridge_generation_modes} gives an example of the two generation modes on miniDomainNet~\cite{zhou2021domain}. The complete training and generation procedure of MUSE is summarized in Appendix~\ref{app:muse_algorithm}.

\section{Experiments}
\label{sec:experiments}

\subsection{Experimental Setup}

\paragraph{Datasets and metrics.}
We evaluate MUSE on three closed-set UDA benchmarks: Office-31~\cite{saenko2010adapting}, Office-Home~\cite{venkateswara2017deep}, and miniDomainNet~\cite{zhou2021domain}.
Office-31 contains Amazon (A), DSLR (D), and Webcam (W), giving six ordered transfer tasks.
Office-Home contains Artistic (Ar), Clipart (Cl), Product (Pr), and Real-World (Rw), giving twelve tasks.
For miniDomainNet, following recent diffusion-based UDA work~\cite{zhang2025dcdm}, we use Clipart (C), Painting (P), Real (R), and Sketch (S), also giving twelve tasks.
All domains in each benchmark share the same label space.
We report target-domain top-1 accuracy (\%) for each task and the average over all tasks. The highest value in each column is shown in bold, and the highest value strictly below it is underlined.

\paragraph{Methods compared.}
Since MUSE adapts a generator rather than a classifier, we combine it with three downstream UDA methods: two CNN-based methods, MCC~\cite{jin2020minimum} and ELS~\cite{zhang2023els}, and one transformer-based method, SSRT~\cite{sun2022ssrt}.
We compare against representative discriminative UDA methods, including DANN~\cite{ganin2016domain}, CDAN~\cite{long2018conditional}, AFN~\cite{xu2019adaptive}, MDD~\cite{zhang2019bridging}, SDAT~\cite{rangwani2022sdat}, MSGD~\cite{xia2023msgd}, MCC, ELS, and SSRT.
For diffusion-based baselines, we use repeated single-target adaptation with Terra~\cite{zhuang2024terra} and DCDM~\cite{zhang2025dcdm}, paired with the same downstream learners when applicable.
These baselines fine-tune a separate generator for each source--target pair, whereas MUSE performs one multi-target generator adaptation process for each source domain.

\paragraph{Generation and downstream protocol.}
MUSE is built on an SDXL-based~\cite{SDXL} latent diffusion backbone, with trainable adaptation restricted to the U-Net attention projections.
For a fair comparison, all diffusion-based methods use the same synthetic-data protocol: for each target domain, we generate 50 class-conditional samples per class and source-to-target translated samples via DDIM inversion, and use their union as synthetic supervision for downstream UDA.
MCC, ELS, and SSRT follow their original training recipes and backbone settings~\cite{jin2020minimum,zhang2023els,sun2022ssrt}.
Implementation details are provided in Appendix~\ref{app:implementation_details}.

\begin{table*}[t]
\centering
\footnotesize
\setlength{\tabcolsep}{3.5pt}
\renewcommand{\arraystretch}{1.03}
\caption{Transfer accuracy (\%) on Office-31. Avg. is over six tasks. Pairwise denotes per-source--target diffusion generation; Multi-target denotes one source-guided generator reused across targets.}
\label{tab:office31_main}
\begin{tabular}{lcccccccc}
\toprule
Method 
& \shortstack{Diffusion Gen.\\Protocol}
& A$\rightarrow$W & D$\rightarrow$W & W$\rightarrow$D 
& A$\rightarrow$D & D$\rightarrow$A & W$\rightarrow$A & Avg. \\
\midrule
ERM & \textemdash & 77.07 & 96.60 & 99.20 & 81.08 & 64.11 & 64.01 & 80.35 \\
DANN~\cite{ganin2016domain} & \textemdash & 89.85 & 97.95 & 99.90 & 83.26 & 73.28 & 73.75 & 86.33 \\
CDAN~\cite{long2018conditional} & \textemdash & 92.42 & 98.62 & \textbf{100.00} & 91.44 & 74.61 & 72.80 & 88.32 \\
AFN~\cite{xu2019adaptive} & \textemdash & 91.82 & 98.77 & \textbf{100.00} & 95.12 & 72.43 & 70.71 & 88.14 \\
MDD~\cite{zhang2019bridging} & \textemdash & 93.55 & 98.66 & \textbf{100.00} & 93.92 & 75.29 & 73.95 & 89.23 \\
SDAT~\cite{rangwani2022sdat} & \textemdash & 91.32 & 98.83 & \textbf{100.00} & 95.25 & 76.97 & 73.19 & 89.26 \\
MSGD~\cite{xia2023msgd} & \textemdash & 95.50 & 99.20 & \textbf{100.00} & 95.60 & 77.30 & 77.00 & 90.80 \\
\rowcolor{mccbg}
MCC~\cite{jin2020minimum} & \textemdash & 94.09 & 98.32 & 99.67 & 94.25 & 75.89 & 75.46 & 89.61 \\
\rowcolor{elsbg}
ELS~\cite{zhang2023els} & \textemdash & 93.84 & 98.78 & \textbf{100.00} & 95.78 & 77.72 & 75.13 & 90.21 \\
\rowcolor{ssrtbg}
SSRT~\cite{sun2022ssrt} & \textemdash & 97.70 & 99.20 & \textbf{100.00} & 98.60 & 83.50 & 82.20 & 93.50 \\
\midrule
\multicolumn{9}{l}{Diffusion-based baselines: pairwise generation} \\
\cmidrule(lr){1-9}
\rowcolor{mccbg}
MCC+Terra~\cite{zhuang2024terra} & Pairwise & 94.55 & 99.03 & \textbf{100.00} & 96.46 & 78.64 & 79.37 & 91.34 \\
\rowcolor{elsbg}
ELS+Terra~\cite{zhuang2024terra} & Pairwise & 94.09 & 99.21 & \textbf{100.00} & 96.25 & 78.67 & 79.45 & 91.28 \\
\rowcolor{ssrtbg}
SSRT+Terra~\cite{zhuang2024terra} & Pairwise & 98.19 & 98.49 & \textbf{100.00} & 98.19 & 83.82 & 83.71 & 93.73 \\
\rowcolor{mccbg}
MCC+DCDM~\cite{zhang2025dcdm} & Pairwise & 95.51 & 98.58 & \underline{99.93} & 95.31 & 78.26 & 78.43 & 91.01 \\
\rowcolor{elsbg}
ELS+DCDM~\cite{zhang2025dcdm} & Pairwise & 96.90 & 98.91 & \textbf{100.00} & 97.46 & 79.79 & 77.74 & 91.80 \\
\rowcolor{ssrtbg}
SSRT+DCDM~\cite{zhang2025dcdm} & Pairwise & \textbf{100.00} & \textbf{100.00} & \textbf{100.00} & \textbf{99.90} & \underline{85.06} & \underline{86.69} & \underline{95.29} \\
\midrule
\multicolumn{9}{l}{\textbf{Our method: multi-target generation}} \\
\cmidrule(lr){1-9}
\rowcolor{mccbg}
MCC+MUSE & Multi-target & 94.52 & 98.86 & \textbf{100.00} & 96.45 & 80.25 & 80.52 & 91.77 \\
\rowcolor{elsbg}
ELS+MUSE & Multi-target & 95.01 & \underline{99.43} & \textbf{100.00} & 96.86 & 81.81 & 81.06 & 92.36 \\
\rowcolor{ssrtbg}
SSRT+MUSE & Multi-target & \underline{99.92} & \textbf{100.00} & \textbf{100.00} & \underline{99.61} & \textbf{86.62} & \textbf{87.02} & \textbf{95.53} \\
\bottomrule
\end{tabular}
\end{table*}

\begin{table*}[t]
\centering
\scriptsize
\setlength{\tabcolsep}{3.0pt}
\renewcommand{\arraystretch}{1.06}
\caption{Transfer accuracy (\%) on Office-Home. Avg. is over twelve tasks. Pairwise denotes per-source--target diffusion generation; Multi-target denotes one source-guided generator reused across targets.}
\label{tab:officehome_main}
\resizebox{\textwidth}{!}{%
\begin{tabular}{lcccccccccccccc}
\toprule
Method
& \shortstack{Diffusion Gen.\\Protocol}
& Ar$\rightarrow$Cl & Ar$\rightarrow$Pr & Ar$\rightarrow$Rw
& Cl$\rightarrow$Ar & Cl$\rightarrow$Pr & Cl$\rightarrow$Rw
& Pr$\rightarrow$Ar & Pr$\rightarrow$Cl & Pr$\rightarrow$Rw
& Rw$\rightarrow$Ar & Rw$\rightarrow$Cl & Rw$\rightarrow$Pr
& Avg. \\
\midrule
ERM
& \textemdash & 44.06 & 67.12 & 74.26 & 53.26 & 61.96 & 64.54 & 51.91 & 38.90 & 72.94 & 64.51 & 43.84 & 75.39 & 59.39 \\
DANN~\cite{ganin2016domain}
& \textemdash & 52.53 & 62.57 & 73.20 & 56.89 & 67.02 & 68.34 & 58.37 & 54.14 & 78.31 & 70.78 & 60.76 & 80.57 & 65.29 \\
CDAN~\cite{long2018conditional}
& \textemdash & 54.21 & 72.18 & 78.29 & 61.97 & 71.43 & 72.39 & 62.96 & 55.68 & 80.68 & 74.71 & 61.22 & 83.68 & 69.12 \\
AFN~\cite{xu2019adaptive}
& \textemdash & 52.58 & 72.42 & 76.96 & 64.90 & 71.14 & 72.91 & 64.08 & 51.29 & 77.83 & 72.21 & 57.46 & 82.09 & 67.99 \\
MDD~\cite{zhang2019bridging}
& \textemdash & 56.37 & 75.53 & 79.17 & 62.95 & 73.21 & 73.55 & 62.56 & 54.86 & 79.49 & 73.84 & 61.45 & 84.06 & 69.75 \\
SDAT~\cite{rangwani2022sdat}
& \textemdash & 58.20 & 77.46 & 81.35 & 66.06 & 76.45 & 76.41 & 63.70 & 56.69 & 82.49 & 76.02 & 62.09 & 85.24 & 71.85 \\
MSGD~\cite{xia2023msgd}
& \textemdash & 58.70 & 76.90 & 78.90 & 70.10 & 76.20 & 76.60 & 69.00 & 57.20 & 82.30 & 74.90 & 62.70 & 84.50 & 72.40 \\
\rowcolor{mccbg}
MCC~\cite{jin2020minimum}
& \textemdash & 56.83 & 79.81 & 82.66 & 67.80 & 77.02 & 77.82 & 66.98 & 55.43 & 81.79 & 73.95 & 61.41 & 85.44 & 72.24 \\
\rowcolor{elsbg}
ELS~\cite{zhang2023els}
& \textemdash & 57.79 & 77.65 & 81.62 & 66.59 & 76.74 & 76.43 & 62.69 & 56.69 & 82.12 & 75.63 & 62.85 & 85.35 & 71.84 \\
\rowcolor{ssrtbg}
SSRT~\cite{sun2022ssrt}
& \textemdash & 75.17 & 88.98 & 91.09 & 85.13 & 88.29 & 89.95 & 85.04 & 74.23 & 91.26 & \underline{85.70} & 78.58 & 91.78 & 85.43 \\
\midrule
\multicolumn{15}{l}{Diffusion-based baselines: pairwise generation} \\
\cmidrule(lr){1-15}
\rowcolor{mccbg}
MCC+Terra~\cite{zhuang2024terra}
& Pairwise & 63.49 & 81.51 & 83.46 & 72.52 & 82.89 & 81.25 & 73.20 & 61.66 & 83.16 & 74.36 & 63.45 & 84.41 & 75.45 \\
\rowcolor{elsbg}
ELS+Terra~\cite{zhuang2024terra}
& Pairwise & 64.62 & 82.33 & 83.60 & 71.19 & 84.25 & 80.31 & 73.00 & 63.57 & 83.81 & 76.20 & 66.56 & 85.70 & 76.26 \\
\rowcolor{ssrtbg}
SSRT+Terra~\cite{zhuang2024terra}
& Pairwise & 73.17 & 89.52 & 90.45 & 85.99 & 88.87 & \underline{91.14} & \underline{86.44} & 74.27 & \underline{91.48} & 85.58 & 78.25 & 91.85 & 85.58 \\
\rowcolor{mccbg}
MCC+DCDM~\cite{zhang2025dcdm}
& Pairwise & 58.23 & 80.33 & 82.91 & 70.14 & 79.15 & 81.36 & 68.49 & 57.75 & 83.44 & 74.18 & 63.81 & 85.67 & 73.71 \\
\rowcolor{elsbg}
ELS+DCDM~\cite{zhang2025dcdm}
& Pairwise & 60.35 & 78.81 & 82.74 & 69.59 & 80.53 & 79.55 & 65.16 & 58.26 & 83.11 & 75.81 & 64.18 & 85.55 & 73.64 \\
\rowcolor{ssrtbg}
SSRT+DCDM~\cite{zhang2025dcdm}
& Pairwise & \textbf{79.34} & \textbf{91.55} & \underline{91.65} & \underline{86.53} & \underline{90.04} & 90.38 & 84.96 & \underline{76.59} & 91.39 & 84.47 & \underline{78.92} & \underline{92.70} & \underline{86.54} \\
\midrule
\multicolumn{15}{l}{\textbf{Our method: multi-target generation}} \\
\cmidrule(lr){1-15}
\rowcolor{mccbg}
MCC+MUSE
& Multi-target & 64.31 & 83.44 & 82.37 & 73.14 & 84.44 & 83.50 & 72.90 & 63.02 & 83.47 & 75.37 & 66.58 & 85.82 & 76.53 \\
\rowcolor{elsbg}
ELS+MUSE
& Multi-target & 66.51 & 84.09 & 82.62 & 74.14 & 83.74 & 82.96 & 74.61 & 64.73 & 85.52 & 75.61 & 68.81 & 87.45 & 77.57 \\
\rowcolor{ssrtbg}
SSRT+MUSE
& Multi-target & \underline{79.09} & \underline{90.43} & \textbf{92.44} & \textbf{87.81} & \textbf{92.66} & \textbf{92.69} & \textbf{87.44} & \textbf{79.42} & \textbf{93.45} & \textbf{88.52} & \textbf{80.79} & \textbf{93.01} & \textbf{88.15} \\
\bottomrule
\end{tabular}%
}
\end{table*}

\begin{table*}[t]
\centering
\scriptsize
\setlength{\tabcolsep}{3.0pt}
\renewcommand{\arraystretch}{1.06}
\caption{Transfer accuracy (\%) on miniDomainNet. Avg. is over twelve tasks. Pairwise denotes per-source--target diffusion generation; Multi-target denotes one source-guided generator reused across targets.}
\label{tab:minidomainnet_main}
\resizebox{\textwidth}{!}{%
\begin{tabular}{lcccccccccccccc}
\toprule
Method
& \shortstack{Diffusion Gen.\\Protocol}
& C$\rightarrow$P & C$\rightarrow$R & C$\rightarrow$S
& P$\rightarrow$C & P$\rightarrow$R & P$\rightarrow$S
& R$\rightarrow$C & R$\rightarrow$P & R$\rightarrow$S
& S$\rightarrow$C & S$\rightarrow$P & S$\rightarrow$R
& Avg. \\
\midrule
ERM
& \textemdash & 39.48 & 53.27 & 42.93 & 49.55 & 68.18 & 41.99 & 49.30 & 55.52 & 37.35 & 54.60 & 45.33 & 53.08 & 49.22 \\
DANN~\cite{ganin2016domain}
& \textemdash & 45.94 & 56.12 & 49.40 & 50.72 & 65.61 & 50.07 & 55.15 & 60.55 & 49.95 & 58.54 & 54.64 & 58.99 & 54.64 \\
AFN~\cite{xu2019adaptive}
& \textemdash & 49.23 & 60.11 & 51.11 & 55.60 & 70.59 & 51.78 & 55.84 & 60.41 & 47.46 & 60.69 & 56.36 & 62.28 & 56.79 \\
CDAN~\cite{long2018conditional}
& \textemdash & 47.99 & 58.50 & 51.17 & 56.36 & 68.71 & 53.01 & 61.15 & 62.85 & 53.44 & 60.89 & 55.90 & 60.88 & 57.57 \\
MDD~\cite{zhang2019bridging}
& \textemdash & 48.53 & 61.75 & 52.32 & 59.74 & 70.62 & 55.43 & 62.18 & 62.22 & 54.04 & 63.07 & 58.55 & 64.50 & 59.41 \\
SDAT~\cite{rangwani2022sdat}
& \textemdash & 50.97 & 62.42 & 53.91 & 60.57 & 69.97 & 55.85 & 64.39 & 64.83 & 55.86 & 64.07 & 59.43 & 64.28 & 60.55 \\
\rowcolor{mccbg}
MCC~\cite{jin2020minimum}
& \textemdash & 51.95 & 67.73 & 52.66 & 60.96 & 76.61 & 54.67 & 64.15 & 64.02 & 50.34 & 63.64 & 59.68 & 69.78 & 61.35 \\
\rowcolor{elsbg}
ELS~\cite{zhang2023els}
& \textemdash & 50.11 & 61.45 & 53.02 & 60.77 & 70.61 & 56.04 & 62.43 & 64.16 & 54.89 & 63.93 & 59.19 & 64.47 & 60.09 \\
\rowcolor{ssrtbg}
SSRT~\cite{sun2022ssrt}
& \textemdash & 77.30 & 86.35 & 75.24 & 80.79 & 88.10 & 74.92 & \underline{84.29} & 80.16 & 78.10 & 81.75 & 75.71 & 86.67 & 80.78 \\
\midrule
\multicolumn{15}{l}{Diffusion-based baselines: pairwise generation} \\
\cmidrule(lr){1-15}
\rowcolor{mccbg}
MCC+Terra~\cite{zhuang2024terra}
& Pairwise & 52.64 & 68.26 & 51.88 & 60.45 & 77.24 & 55.41 & 63.85 & 64.52 & 50.86 & 63.21 & 59.88 & 70.68 & 61.57 \\
\rowcolor{elsbg}
ELS+Terra~\cite{zhuang2024terra}
& Pairwise & 52.68 & 68.57 & 54.37 & 60.23 & 70.78 & 56.87 & 63.57 & 63.25 & 54.51 & 64.22 & 60.36 & 64.54 & 61.16 \\
\rowcolor{ssrtbg}
SSRT+Terra~\cite{zhuang2024terra}
& Pairwise & \underline{79.84} & \underline{89.05} & 76.35 & 80.48 & \underline{88.41} & 75.56 & 82.84 & 78.25 & \underline{78.40} & 80.95 & \underline{77.14} & \underline{88.41} & \underline{81.31} \\
\rowcolor{mccbg}
MCC+DCDM~\cite{zhang2025dcdm}
& Pairwise & 55.28 & 70.21 & 54.42 & 63.25 & 77.05 & 55.74 & 65.16 & 65.01 & 51.96 & 65.76 & 60.83 & 71.09 & 62.98 \\
\rowcolor{elsbg}
ELS+DCDM~\cite{zhang2025dcdm}
& Pairwise & 56.14 & 68.52 & 55.51 & 64.96 & 73.44 & 58.23 & 64.67 & 64.36 & 57.66 & 67.23 & 62.56 & 69.90 & 63.60 \\
\rowcolor{ssrtbg}
SSRT+DCDM~\cite{zhang2025dcdm}
& Pairwise & 78.26 & 87.42 & \underline{77.56} & \underline{81.64} & 87.15 & \underline{76.82} & 81.92 & \underline{80.35} & 76.51 & \underline{82.34} & 76.92 & 88.21 & 81.26 \\
\midrule
\multicolumn{15}{l}{\textbf{Our method: multi-target generation}} \\
\cmidrule(lr){1-15}
\rowcolor{mccbg}
MCC+MUSE
& Multi-target & 58.02 & 72.82 & 56.54 & 66.23 & 74.56 & 57.74 & 63.42 & 65.87 & 54.67 & 66.74 & 60.98 & 70.64 & 64.02 \\
\rowcolor{elsbg}
ELS+MUSE
& Multi-target & 59.18 & 73.61 & 57.13 & 65.81 & 75.81 & 58.62 & 64.31 & 65.32 & 58.94 & 67.59 & 61.93 & 72.24 & 65.04 \\
\rowcolor{ssrtbg}
SSRT+MUSE
& Multi-target & \textbf{80.68} & \textbf{91.16} & \textbf{80.05} & \textbf{85.60} & \textbf{91.63} & \textbf{79.25} & \textbf{85.44} & \textbf{81.79} & \textbf{80.05} & \textbf{84.49} & \textbf{80.21} & \textbf{92.11} & \textbf{84.37} \\
\bottomrule
\end{tabular}%
}
\end{table*}

\subsection{Main Results}

Tables~\ref{tab:office31_main}, \ref{tab:officehome_main}, and~\ref{tab:minidomainnet_main} compare MUSE with discriminative UDA methods and repeated per-target diffusion adaptation.
The most direct comparison is against Terra and DCDM under the same downstream UDA method, since this isolates the effect of the synthetic-data generation strategy.
Standard deviations over three runs and qualitative samples are reported in Appendices~\ref{app:standard_deviations} and~\ref{app:qualitative_visualization}.

\paragraph{Office-31.}
MUSE obtains average accuracies of 91.77\%, 92.36\%, and 95.53\% with MCC, ELS, and SSRT, respectively.
All improvements are reported as absolute accuracy gains in percentage points.
Compared with the corresponding discriminative baselines, MUSE improves over MCC, ELS, and SSRT by 2.16, 2.15, and 2.03 points, respectively.
Compared with Terra under the same downstream learners, MUSE improves by 0.43, 1.08, and 1.80 points with MCC, ELS, and SSRT, respectively; compared with DCDM, the corresponding gains are 0.76, 0.56, and 0.24 points.
The gains are most visible on transfers to Amazon, while several other tasks are nearly saturated, especially W$\rightarrow$D.

\paragraph{Office-Home.}
On Office-Home, MUSE obtains average accuracies of 76.53\%, 77.57\%, and 88.15\% with MCC, ELS, and SSRT, respectively.
Compared with the corresponding discriminative baselines, MUSE improves over MCC, ELS, and SSRT by 4.29, 5.73, and 2.72 points, respectively.
Compared with Terra under the same downstream learners, MUSE improves by 1.08, 1.31, and 2.57 points with MCC, ELS, and SSRT, respectively; compared with DCDM, the corresponding gains are 2.82, 3.93, and 1.61 points.
These results show that MUSE consistently improves average performance over both discriminative and diffusion-based baselines across all three downstream learners.
Although DCDM remains stronger on a few individual tasks, MUSE gives the best average performance.

\paragraph{miniDomainNet.}
On miniDomainNet, MUSE obtains average accuracies of 64.02\%, 65.04\%, and 84.37\% with MCC, ELS, and SSRT, respectively.
Compared with the corresponding discriminative baselines, MUSE improves over MCC, ELS, and SSRT by 2.67, 4.95, and 3.59 points, respectively.
Compared with Terra under the same downstream learners, MUSE improves by 2.45, 3.88, and 3.06 points with MCC, ELS, and SSRT, respectively; compared with DCDM, the corresponding gains are 1.04, 1.44, and 3.11 points.
The improvements are largest on this benchmark, where the visual gap among domains is more pronounced.
With SSRT, MUSE achieves the best result on every reported transfer task, suggesting that the shared-private design is particularly effective in visually diverse multi-target settings.

\paragraph{Feature-space visualization.}
To further examine whether the generated samples reduce the source--target discrepancy, we provide t-SNE~\cite{tsne} visualizations of feature distributions in Appendix~\ref{app:tsne_visualization}. The visualizations compare source-domain samples, target-domain samples, DDIM-inversion adapted source samples, and generated target-domain samples on representative categories from Office-31, Office-Home, and miniDomainNet.

\subsection{Efficiency Analysis}

\begin{table}[t]
\centering
\small
\setlength{\tabcolsep}{7pt}
\renewcommand{\arraystretch}{1.10}
\caption{SDXL fine-tuning time on one A100 80GB GPU. The maximum number of fine-tuning steps is 10000 for both Terra and MUSE. \# Domains denotes the total number of domains in the benchmark.}
\label{tab:finetune_time}
\begin{tabular}{lcccc}
\toprule
Dataset & \# Domains & Terra Time (h) $\downarrow$ & MUSE Time (h) $\downarrow$ & Speedup $\uparrow$ \\
\midrule
Office-31      & 3 & \texttt{96.78} & \texttt{51.74} & \texttt{1.87}$\times$ \\
Office-Home    & 4 & \texttt{196.36} & \texttt{80.79} & \texttt{2.43}$\times$ \\
miniDomainNet  & 4 & \texttt{197.68} & \texttt{81.04} & \texttt{2.44}$\times$ \\
\bottomrule
\end{tabular}
\end{table}

Table~\ref{tab:finetune_time} reports the total generator fine-tuning time under the evaluated multi-target generation protocol.
The efficiency comparison in this subsection focuses on methods that adapt a pretrained diffusion foundation model through source--target fine-tuning.
Specifically, Terra~\cite{zhuang2024terra} performs repeated source--target diffusion fine-tuning, requiring a separate adapted generator for each target domain, whereas MUSE performs one source-guided multi-target fine-tuning process for each source domain. DCDM is not included in this fine-tuning-time or adapter-parameter comparison, because it does not fine-tune a pretrained diffusion foundation model such as SDXL.
Therefore, its training cost and parameterization are not directly comparable to PEFT-based diffusion adaptation methods.

By avoiding repeated source-involved fine-tuning for each source--target pair, MUSE reduces the total SDXL fine-tuning time from 96.78 to 51.74 hours on Office-31, from 196.36 to 80.79 hours on Office-Home, and from 197.68 to 81.04 hours on miniDomainNet. 
This corresponds to speedups of 1.87$\times$, 2.43$\times$, and 2.44$\times$, respectively. 
These measurements are intended to isolate the cost of adapting the diffusion generator. 
In addition, Appendix~\ref{app:muse_scaling} shows that, under the same PEFT adapter parameterization, MUSE stores the source-supervised semantic branch and the style projection factors once, and adds only lightweight target-private cores for new targets. 
This reduces adapter-parameter growth relative to repeated independent source--target diffusion adaptation.

\subsection{Ablation Overview}
\label{sec:ablation_overview}

Appendix~\ref{app:ablation_studies} provides Office-Home ablations under the same evaluation protocol.
We study the orthogonality regularizer on target-private cores, freezing the shared semantic branch during target-conditioned updates, adaptive versus uniform target sampling, the two bridge-set components, and prompt-only SDXL generation.
These ablations assess whether each component contributes to the accuracy--efficiency trade-off of MUSE. We discuss additional limitations related to the diffusion-generation cost, and generated-sample quality in Appendix~\ref{sec:limitations}.

\section{Conclusion and Future Works}
\label{sec:conclusion_future_work}

In this paper, we revisited diffusion fine-tuning for UDA in a multi-target setting, where one labeled source domain supports synthetic-data generation for multiple unlabeled target domains. We proposed \textsc{MUSE}, a parameter-efficient framework that separates source-supervised semantic adaptation from target-specific style adaptation through a shared semantic branch and lightweight target-private style cores. With branch-decoupled optimization, adaptive target sampling, and bridge generation based on both DDIM inversion and class-conditional synthesis, MUSE avoids repeated source-involved fine-tuning for each source--target pair. Experiments on Office-31, Office-Home, and miniDomainNet show that MUSE improves the accuracy--efficiency trade-off over repeated diffusion adaptation, reducing generator fine-tuning cost while improving average target-domain accuracy.

Future work may extend MUSE beyond the closed-set image classification setting considered here. Promising directions include partial-set, open-set, and label-shifted multi-target adaptation; adaptive layer-wise or rank-wise sharing between semantic and style branches; and applications to multimodal domain adaptation.


\bibliographystyle{plainnat}
\bibliography{references}

\clearpage
\appendix

\section{Parameter Scaling of MUSE in the Multi-Target Setting}
\label{app:muse_scaling}

This appendix compares the number of \emph{trainable adapter parameters} used by MUSE and by a repeated independent single-target reference under the adapter parameterizations defined below. Throughout, we count only the trainable parameters introduced during fine-tuning and exclude the frozen backbone parameters shared by all methods.

\subsection{Setup}

Let \(\mathcal{L}\) denote the set of adapted linear layers in the diffusion backbone. For each layer \(\ell\in\mathcal{L}\), let
\[
W_\ell
\in
\mathbb{R}^{d_\ell^{\mathrm{out}}\times d_\ell^{\mathrm{in}}}
\]
be the frozen pre-trained weight matrix. Define
\begin{equation}
L := |\mathcal{L}|,
\qquad
S :=
\sum_{\ell\in\mathcal{L}}
\bigl(
d_\ell^{\mathrm{in}}
+
d_\ell^{\mathrm{out}}
\bigr).
\label{eq:appendix_LS_def}
\end{equation}
We consider \(M\ge 2\) target domains and allow the shared semantic branch and target-conditioned style branch to use possibly different ranks,
\[
r_{\mathrm{sem}}>0,
\qquad
r_{\mathrm{sty}}>0.
\]

\paragraph{Repeated independent single-target reference.}
As a reference point, consider adapting one diffusion model per target domain, without sharing trainable adapter parameters across targets. For each target domain \(m\in\{1,\dots,M\}\) and each adapted layer \(\ell\), the update is
\begin{equation}
\Delta W_{\ell}^{(m)}
=
U_{\ell,\mathrm{sem}}^{(m)}
V_{\ell,\mathrm{sem}}^{(m)}
+
U_{\ell,\mathrm{sty}}^{(m)}
\Omega_{\ell}^{(m)}
V_{\ell,\mathrm{sty}}^{(m)} ,
\label{eq:independent_baseline}
\end{equation}
where
\[
U_{\ell,\mathrm{sem}}^{(m)}
\in
\mathbb{R}^{d_\ell^{\mathrm{out}}\times r_{\mathrm{sem}}},
\qquad
V_{\ell,\mathrm{sem}}^{(m)}
\in
\mathbb{R}^{r_{\mathrm{sem}}\times d_\ell^{\mathrm{in}}},
\]
and
\[
U_{\ell,\mathrm{sty}}^{(m)}
\in
\mathbb{R}^{d_\ell^{\mathrm{out}}\times r_{\mathrm{sty}}},
\qquad
V_{\ell,\mathrm{sty}}^{(m)}
\in
\mathbb{R}^{r_{\mathrm{sty}}\times d_\ell^{\mathrm{in}}},
\qquad
\Omega_{\ell}^{(m)}
\in
\mathbb{R}^{r_{\mathrm{sty}}\times r_{\mathrm{sty}}}.
\]
For one target and one adapted layer, the number of trainable parameters is
\[
r_{\mathrm{sem}}
\bigl(d_\ell^{\mathrm{in}}+d_\ell^{\mathrm{out}}\bigr)
+
r_{\mathrm{sty}}
\bigl(d_\ell^{\mathrm{in}}+d_\ell^{\mathrm{out}}\bigr)
+
r_{\mathrm{sty}}^2 .
\]
Summing over all adapted layers and all \(M\) targets gives
\begin{equation}
P_{\mathrm{ind}}(M)
=
M
\Bigl[
(r_{\mathrm{sem}}+r_{\mathrm{sty}})S
+
Lr_{\mathrm{sty}}^2
\Bigr].
\label{eq:independent_parameter_count}
\end{equation}

\paragraph{MUSE.}
In MUSE, for each adapted layer \(\ell\) and target domain \(m\), the update is
\begin{equation}
\Delta W_{\ell}^{(m)}
=
U_{\ell,\mathrm{sem}}
V_{\ell,\mathrm{sem}}
+
U_{\ell,\mathrm{sty}}
\Omega_{\ell}^{(m)}
V_{\ell,\mathrm{sty}} ,
\label{eq:muse_appendix_form}
\end{equation}
where
\[
U_{\ell,\mathrm{sem}}
\in
\mathbb{R}^{d_\ell^{\mathrm{out}}\times r_{\mathrm{sem}}},
\qquad
V_{\ell,\mathrm{sem}}
\in
\mathbb{R}^{r_{\mathrm{sem}}\times d_\ell^{\mathrm{in}}},
\]
and
\[
U_{\ell,\mathrm{sty}}
\in
\mathbb{R}^{d_\ell^{\mathrm{out}}\times r_{\mathrm{sty}}},
\qquad
V_{\ell,\mathrm{sty}}
\in
\mathbb{R}^{r_{\mathrm{sty}}\times d_\ell^{\mathrm{in}}}
\]
are shared across all target domains, while
\[
\Omega_{\ell}^{(m)}
\in
\mathbb{R}^{r_{\mathrm{sty}}\times r_{\mathrm{sty}}}
\]
is target-specific. Hence, for one adapted layer, MUSE stores
\[
(r_{\mathrm{sem}}+r_{\mathrm{sty}})
\bigl(d_\ell^{\mathrm{in}}+d_\ell^{\mathrm{out}}\bigr)
+
Mr_{\mathrm{sty}}^2
\]
trainable parameters. Summing over all adapted layers gives
\begin{equation}
P_{\mathrm{MUSE}}(M)
=
(r_{\mathrm{sem}}+r_{\mathrm{sty}})S
+
MLr_{\mathrm{sty}}^2 .
\label{eq:muse_parameter_count}
\end{equation}

\subsection{Main Result}

\paragraph{Proposition.}
Under the parameterizations in Eqs.~\eqref{eq:independent_baseline} and~\eqref{eq:muse_appendix_form}, for every \(M\ge 2\),
\begin{equation}
P_{\mathrm{ind}}(M)
-
P_{\mathrm{MUSE}}(M)
=
(M-1)(r_{\mathrm{sem}}+r_{\mathrm{sty}})S .
\label{eq:parameter_count_difference}
\end{equation}
In particular, MUSE uses strictly fewer trainable adapter parameters than this repeated independent reference for every \(M\ge 2\).

\paragraph{Proof.}
Substituting Eqs.~\eqref{eq:independent_parameter_count} and~\eqref{eq:muse_parameter_count} yields
\begin{align*}
P_{\mathrm{ind}}(M)
-
P_{\mathrm{MUSE}}(M)
&=
M
\Bigl[
(r_{\mathrm{sem}}+r_{\mathrm{sty}})S
+
Lr_{\mathrm{sty}}^2
\Bigr]
-
\Bigl[
(r_{\mathrm{sem}}+r_{\mathrm{sty}})S
+
MLr_{\mathrm{sty}}^2
\Bigr]
\\
&=
(M-1)(r_{\mathrm{sem}}+r_{\mathrm{sty}})S .
\end{align*}
Since \(M\ge 2\), \(r_{\mathrm{sem}}>0\), \(r_{\mathrm{sty}}>0\), and \(S>0\), the right-hand side is strictly positive.
\hfill \(\square\)

\paragraph{Corollary.}
The incremental parameter cost of adding one target domain is
\begin{equation}
\begin{aligned}
P_{\mathrm{ind}}(M+1)-P_{\mathrm{ind}}(M)
&=
(r_{\mathrm{sem}}+r_{\mathrm{sty}})S
+
Lr_{\mathrm{sty}}^2,
\\
P_{\mathrm{MUSE}}(M+1)-P_{\mathrm{MUSE}}(M)
&=
Lr_{\mathrm{sty}}^2 .
\end{aligned}
\label{eq:incremental_parameter_cost}
\end{equation}

\paragraph{Proof.}
Both identities follow directly from Eqs.~\eqref{eq:independent_parameter_count} and~\eqref{eq:muse_parameter_count}.
\hfill \(\square\)

\paragraph{Equal-rank special case.}
If \(r_{\mathrm{sem}}=r_{\mathrm{sty}}=r\), then
\[
P_{\mathrm{ind}}(M)
=
M(2rS+Lr^2),
\qquad
P_{\mathrm{MUSE}}(M)
=
2rS+MLr^2,
\]
and Eq.~\eqref{eq:parameter_count_difference} reduces to
\[
P_{\mathrm{ind}}(M)
-
P_{\mathrm{MUSE}}(M)
=
2(M-1)rS .
\]

\paragraph{Remark.}
This comparison isolates the effect of cross-target sharing in the specified adapter parameterization. In the repeated independent reference, each target stores its own semantic factors and target-conditioned style factors. In MUSE, the semantic branch and the style projection factors are stored once and reused across targets, while only the cores
\(\{\Omega_{\ell}^{(m)}\}_{m=1}^{M}\) remain target-specific. Equation~\eqref{eq:parameter_count_difference} quantifies the resulting reduction in trainable adapter parameters, and Eq.~\eqref{eq:incremental_parameter_cost} gives the corresponding per-target incremental cost.

\subsection{Memory and Training-Time Scaling}
\label{app:memory_time_scaling}

We further compare the memory and training-time requirements of MUSE with repeated independent diffusion fine-tuning using Terra~\cite{zhuang2024terra}. Peak-memory measurements use the same SDXL backbone, image resolution, batch size, numerical precision, gradient checkpointing, memory-efficient attention, and optimizer settings. We report the peak allocated VRAM of each training process, defined as the maximum memory occupied by PyTorch-allocated tensors during training. The cumulative fine-tuning times correspond to the complete multi-target generation protocol reported in the main paper.

For a fixed source domain, \(M\) denotes the number of target domains that must be supported. Independent fine-tuning therefore requires \(M\) separate source--target training processes, whereas MUSE handles all \(M\) targets within one source-guided multi-target training process.

\begin{table*}[t]
\centering
\scriptsize
\setlength{\tabcolsep}{4.0pt}
\renewcommand{\arraystretch}{1.08}
\caption{Memory and training-time comparison for multi-target diffusion adaptation. Peak allocated VRAM is measured per training process. Aggregate VRAM for independent fine-tuning corresponds to concurrently executing the \(M\) target-specific processes and is obtained by summing their unrounded per-process peaks. Cumulative fine-tuning time is measured over the complete benchmark protocol.}
\label{tab:memory_time_scaling}
\resizebox{\textwidth}{!}{%
\begin{tabular}{llcccc}
\toprule
Dataset
& Method
& \(M\)
& \shortstack{Peak allocated VRAM\\per process (GiB)}
& \shortstack{Aggregate VRAM across\\\(M\) concurrent processes (GiB)}
& \shortstack{Cumulative\\fine-tuning time (h)} \\
\midrule
Office-31
& Independent fine-tuning (Terra)
& 2
& 32.433
& 64.865
& 96.78 \\
Office-31
& MUSE
& 2
& 27.849
& 27.849
& 51.74 \\
\midrule
Office-Home
& Independent fine-tuning (Terra)
& 3
& 32.433
& 97.298
& 196.36 \\
Office-Home
& MUSE
& 3
& 27.835
& 27.835
& 80.79 \\
\midrule
miniDomainNet
& Independent fine-tuning (Terra)
& 3
& 32.433
& 97.298
& 197.68 \\
miniDomainNet
& MUSE
& 3
& 27.835
& 27.835
& 81.04 \\
\bottomrule
\end{tabular}%
}
\end{table*}

The aggregate values for independent fine-tuning in Table~\ref{tab:memory_time_scaling} represent the memory required when the \(M\) target-specific training processes are executed concurrently; they are obtained by summing the unrounded peak allocated VRAM values of the corresponding single-target processes and are not measurements from a single process. If the independent fine-tunings are instead executed sequentially, their peak VRAM is that of one Terra process, while the cumulative training time still grows with the number of source--target runs.

Under the complete evaluation protocol, Terra requires 96.78, 196.36, and 197.68 hours on Office-31, Office-Home, and miniDomainNet, respectively, whereas MUSE requires 51.74, 80.79, and 81.04 hours. When the \(M\) independent target-specific runs are executed concurrently, their aggregate memory grows approximately linearly with \(M\), whereas MUSE uses a single joint training process. MUSE also has a lower peak allocated VRAM than one Terra process. Moreover, its peak memory remains nearly unchanged from \(M=2\) to \(M=3\), because each target-conditioned denoising pass activates only the sampled target branch rather than simultaneously evaluating all target-specific branches. Overall, MUSE reduces cumulative fine-tuning time under sequential execution while avoiding the aggregate-memory growth associated with concurrently running \(M\) independent target-specific fine-tunings.

\section{Complete MUSE Training and Generation Algorithm}
\label{app:muse_algorithm}

Algorithms~\ref{alg:muse_training} and~\ref{alg:muse_generation} summarize the complete MUSE procedure, including branch-decoupled diffusion adaptation, adaptive target-domain sampling, target-specific bridge generation, and downstream UDA training. Let
\[
\Phi_{\mathrm{sem}}
=
\{U_{\ell,\mathrm{sem}},V_{\ell,\mathrm{sem}}\}_{\ell\in\mathcal{L}}
\]
denote the shared semantic parameters, and let
\[
\Phi_{\mathrm{sty}}
=
\{U_{\ell,\mathrm{sty}},V_{\ell,\mathrm{sty}},
\Omega_{\ell}^{(1)},\ldots,\Omega_{\ell}^{(M)}\}_{\ell\in\mathcal{L}}
\]
denote the target-conditioned style parameters. For brevity, we write
\[
\Omega^{(m)}
=
\{\Omega_{\ell}^{(m)}\}_{\ell\in\mathcal{L}}
\]
for the collection of target-private cores associated with target domain \(m\).

\begin{algorithm}[t]
\footnotesize
\caption{MUSE Diffusion Adaptation}
\label{alg:muse_training}
\begin{algorithmic}[1]

\Require Labeled source domain
\(\mathcal{D}_s=\{(x_i^s,y_i^s)\}_{i=1}^{N_s}\);
unlabeled target domains
\(\{\mathcal{D}_t^{(m)}\}_{m=1}^{M}\);
frozen diffusion backbone \(\theta_0\)

\Ensure Adapted shared semantic parameters
\(\Phi_{\mathrm{sem}}\)
and target-conditioned style parameters
\(\Phi_{\mathrm{sty}}\)

\State Initialize \(\Phi_{\mathrm{sem}}\) and \(\Phi_{\mathrm{sty}}\)
\State Initialize target-loss EMAs
\(\{\bar{\ell}_m\}_{m=1}^{M}\) to the same value
and set \(p_m\gets 1/M\) for all \(m\)

\For{each training iteration \(b\)}

    \Statex
    \State \textbf{Source-conditioned pass}
    \State Sample a labeled source minibatch
    \(\{(x_i^s,y_i^s)\}\sim\mathcal{D}_s\)
    \State Deactivate all target-private cores
    \(\{\Omega^{(m)}\}_{m=1}^{M}\)
    \State Compute the source objective
    \(\widehat{\mathcal{J}}_{\mathrm{src}}\)
    using class prompts
    \(p_s(y_i^s)=\texttt{``a }y_i^s\texttt{''}\)
    \State Backpropagate
    \(\widehat{\mathcal{J}}_{\mathrm{src}}\)
    only through \(\Phi_{\mathrm{sem}}\)

    \Statex
    \State \textbf{Target-domain selection}
    \State Sample
    \(m_b\sim\mathrm{Categorical}(p_1,\ldots,p_M)\)

    \Statex
    \State \textbf{Target-conditioned pass}
    \State Sample an unlabeled target minibatch
    \(\{x_j^{t,m_b}\}\sim\mathcal{D}_t^{(m_b)}\)
    \State Activate only the target-private cores
    \(\Omega^{(m_b)}\)
    \State Use the generic target prompt
    \(p_t=\texttt{``an image''}\)
    and stop gradients through \(\Phi_{\mathrm{sem}}\)
    \State Compute the sampled-target denoising loss
    \(\widehat{\mathcal{L}}_{\mathrm{tgt}}^{(m_b)}\)
    \State Compute
    \(\mathcal{L}_{\mathrm{orth}}\)
    and
    \(\mathcal{L}_{\mathrm{core}}\)
    over all target-private cores
    \State Form
    \[
    \widehat{\mathcal{J}}_{\mathrm{tar}}
    =
    \lambda_{m_b}
    \widehat{\mathcal{L}}_{\mathrm{tgt}}^{(m_b)}
    +
    \lambda_{\mathrm{orth}}
    \mathcal{L}_{\mathrm{orth}}
    +
    \lambda_{\mathrm{core}}
    \mathcal{L}_{\mathrm{core}}
    \]
    \State Backpropagate
    \(\widehat{\mathcal{J}}_{\mathrm{tar}}\)
    only through \(\Phi_{\mathrm{sty}}\)

    \Statex
    \State \textbf{Parameter and sampling-statistic updates}
    \State Apply one optimizer step using gradients from both passes
    \State Update the sampled target's EMA:
    \[
    \bar{\ell}_{m_b}
    \gets
    \beta\bar{\ell}_{m_b}
    +
    (1-\beta)
    \widehat{\mathcal{L}}_{\mathrm{tgt}}^{(m_b)}
    \]
    and keep \(\bar{\ell}_k\) unchanged for \(k\neq m_b\)

    \If{\(b\) is a target-probability update step}
        \State Update, for \(m=1,\ldots,M\),
        \[
        p_m
        \gets
        \frac{(\bar{\ell}_m+\tau)^\alpha}
        {\sum_{k=1}^{M}(\bar{\ell}_k+\tau)^\alpha}
        \]
    \EndIf

\EndFor

\end{algorithmic}
\end{algorithm}

\begin{algorithm}[t]
\footnotesize
\caption{Target-Specific Bridge Generation and Downstream Adaptation}
\label{alg:muse_generation}
\begin{algorithmic}[1]

\Require Adapted parameters
\(\Phi_{\mathrm{sem}}\) and \(\Phi_{\mathrm{sty}}\);
labeled source domain \(\mathcal{D}_s\);
unlabeled target domains
\(\{\mathcal{D}_t^{(m)}\}_{m=1}^{M}\)

\Ensure Target-specific labeled bridge sets
\(\{\widehat{\mathcal{B}}^{(m)}\}_{m=1}^{M}\)
and downstream adapted learners
\(\{f^{(m)}\}_{m=1}^{M}\)

\For{\(m=1,\ldots,M\)}

    \State Reuse the shared semantic branch and shared style projection factors
    and activate only \(\Omega^{(m)}\)

    \State Generate labeled source-to-target translated samples
    \(\widehat{\mathcal{B}}_{\mathrm{inv}}^{(m)}\)
    using DDIM inversion

    \State Generate class-conditional target-style samples
    \(\widehat{\mathcal{B}}_{\mathrm{cls}}^{(m)}\)
    from Gaussian noise using
    \(p_s(y)=\texttt{``a }y\texttt{''}\)

    \State Construct the target-specific bridge set
    \[
    \widehat{\mathcal{B}}^{(m)}
    =
    \widehat{\mathcal{B}}_{\mathrm{inv}}^{(m)}
    \cup
    \widehat{\mathcal{B}}_{\mathrm{cls}}^{(m)}
    \]

    \State Train the downstream UDA learner \(f^{(m)}\) using
    \(\widehat{\mathcal{B}}^{(m)}\)
    and the unlabeled target domain
    \(\mathcal{D}_t^{(m)}\)

\EndFor

\end{algorithmic}
\end{algorithm}

In our implementation, target-domain sampling is initialized uniformly, and the sampling probabilities are recomputed every \(20\) optimization steps using
\(\beta=0.9\), \(\tau=10^{-6}\), and \(\alpha=1\).
The source- and target-side objectives are backpropagated through disjoint parameter subsets before a single optimizer update. During the target-conditioned pass, the shared semantic branch participates in the forward computation but is treated as stop-gradient, while gradients are routed to the shared style projection factors and target-private style cores. Only the private cores associated with the sampled target are activated for the target denoising loss, whereas the core regularizers are evaluated over all target-private cores.

\section{Optimization Details}
\label{app:optimization_details}

\subsection{Two-Pass Gradient Routing Within One Training Iteration}

We summarize the optimization procedure used in the implementation. At iteration \(b\), let
\(\widehat{\mathcal{J}}_{\mathrm{src},b}\) and
\(\widehat{\mathcal{J}}_{\mathrm{tar},b}\) denote the source-side and target-side minibatch objectives defined in Sec.~\ref{sec:method}. The source-conditioned pass deactivates the target-conditioned style term and computes gradients only for the shared semantic branch, while the target-conditioned pass freezes \(\Phi_{\mathrm{sem}}\) and computes gradients only for \(\Phi_{\mathrm{sty}}\):
\begin{equation}
\begin{aligned}
g_{\mathrm{sem}}
&=
\nabla_{\Phi_{\mathrm{sem}}}
\widehat{\mathcal{J}}_{\mathrm{src},b},
\\
g_{\mathrm{sty}}
&=
\nabla_{\Phi_{\mathrm{sty}}}
\widehat{\mathcal{J}}_{\mathrm{tar},b}.
\end{aligned}
\label{eq:two_pass_gradients}
\end{equation}
After both backward passes, the optimizer performs a single update step on the union of trainable parameters. Since the two gradients are routed to disjoint parameter subsets, source-side gradients do not modify the style branch, and target-side gradients do not modify the semantic branch, thereby avoiding direct source--target gradient interference on the shared semantic parameters. Within the target-side pass, the sampled denoising loss activates the sampled target branch, while the core regularizers can also contribute gradients to other target-private cores. Therefore, although each iteration contains two passes, the procedure does not alternately optimize the same parameter subset: the two objectives are backpropagated through disjoint branches before a single optimizer update. This branch-level separation makes the optimization less susceptible to one objective directly overwriting the parameters optimized by the other.

\subsection{Effective Target Weighting Under Adaptive Sampling}

Adaptive target sampling changes the effective weighting of target-domain denoising losses during training. In our implementation, the sampled target loss is not importance-corrected, because adaptive sampling is intended to allocate more optimization effort to domains with larger recent denoising losses rather than to estimate a fixed uniformly weighted target objective.

To see this, fix the current sampling distribution \(p=(p_1,\dots,p_M)\). A target minibatch is constructed by first sampling a domain index \(m\sim p\), then sampling an image \(x^{t,m}\sim\mathcal{D}_t^{(m)}\), and finally sampling the diffusion variables \(t\) and \(\epsilon\). Let \(\eta^{t,m}=\eta(z_0^{t,m},t,\epsilon)\). Ignoring regularization terms, the expected denoising loss of one sampled target example is
\begin{align}
&\mathbb{E}_{m\sim p}
\Bigl[
\lambda_m
\mathbb{E}_{x^{t,m},t,\epsilon}
\bigl[
w(t)
\bigl\|
\eta^{t,m}-
D_{\theta_0,\phi}(z_t^{t,m},t,p_t,m)
\bigr\|_2^2
\bigr]
\Bigr]
\nonumber\\
&\qquad =
\sum_{m=1}^{M}
p_m\lambda_m
\mathcal{L}_{\mathrm{sty}}^{(m)} .
\label{eq:effective_weighting}
\end{align}
Thus, conditioned on the current sampling distribution \(p\), the stochastic target denoising loss is an unbiased estimator of the probability-weighted denoising objective
\begin{equation}
\mathcal{J}_{\mathrm{tar}}
(\Phi_{\mathrm{sty}};\Phi_{\mathrm{sem}},p)
=
\sum_{m=1}^{M}
p_m\lambda_m
\mathcal{L}_{\mathrm{sty}}^{(m)}
+
\lambda_{\mathrm{orth}}\mathcal{L}_{\mathrm{orth}}
+
\lambda_{\mathrm{core}}\mathcal{L}_{\mathrm{core}} .
\label{eq:effective_target_objective}
\end{equation}
This differs from estimating the fixed target denoising objective
\(\sum_{m=1}^{M}\lambda_m\mathcal{L}_{\mathrm{sty}}^{(m)}\), which would require importance correction by \(1/p_m\) for the sampled domain \(m\).

Since \(p_m\) is updated online using exponential moving averages of the observed target losses, the effective domain weights \(p_m\lambda_m\) vary over training. Adaptive sampling therefore acts as a dynamic domain-reweighting mechanism for the target denoising term: target domains with persistently larger denoising losses are sampled more frequently and consequently receive more denoising updates. The core regularizers are evaluated over all target-private cores and are not sampled in the same way.

\subsection{Interpretation of the Core Regularizers}

The regularizers in Eq.~\eqref{eq:reg_terms_main} are applied to the target-private cores
\(\{\Omega_\ell^{(m)}\}\). The orthogonality regularizer
\[
\mathcal{L}_{\mathrm{orth}}
=
\sum_{\ell}
\sum_{1\le i<j\le M}
\left\|
(\Omega_\ell^{(i)})^\top
\Omega_\ell^{(j)}
\right\|_F^2
\]
penalizes highly correlated core coefficients across different target domains at the same layer. Since the induced target-specific update is
\[
\Delta W_{\ell,\mathrm{sty}}^{(m)}
=
U_{\ell,\mathrm{sty}}
\Omega_\ell^{(m)}
V_{\ell,\mathrm{sty}},
\]
this penalty encourages different targets to select distinct coefficient patterns within the shared style subspace parameterized by
\(U_{\ell,\mathrm{sty}}\) and \(V_{\ell,\mathrm{sty}}\).

The magnitude regularizer
\[
\mathcal{L}_{\mathrm{core}}
=
\sum_{\ell}
\sum_{m=1}^{M}
\|\Omega_\ell^{(m)}\|_F^2
\]
controls the scale of the target-private cores, thereby limiting overly large target-conditioned deviations from the frozen diffusion backbone.

Because \(U_{\ell,\mathrm{sty}}\) and \(V_{\ell,\mathrm{sty}}\) are learned without explicit orthonormality constraints, these penalties should be interpreted as lightweight coefficient-space regularization on the target-private cores. They encourage small and diverse core coefficients, but do not impose exact constraints on the magnitude or mutual orthogonality of the induced full updates
\(U_{\ell,\mathrm{sty}}\Omega_\ell^{(m)}V_{\ell,\mathrm{sty}}\).

\section{Implementation Details}
\label{app:implementation_details}

This section provides the implementation details used in our experiments. Unless otherwise specified, all hyper-parameters are kept fixed across runs.

\paragraph{Diffusion backbone.}
We implement MUSE on top of the \textsc{SDXL} backbone~\cite{SDXL}. The VAE is kept in fp32 during training, while the diffusion U-Net forward and backward passes are performed with fp16 mixed precision. The VAE encoder, both text encoders, and all original U-Net weights are frozen throughout training. Only the newly introduced parameter-efficient adapter parameters in the U-Net attention projections are optimized.

\paragraph{Data preprocessing.}
Each image is converted to RGB, resized to resolution \(1024\times1024\), randomly cropped, and normalized to \([-1,1]\) with channel-wise mean and standard deviation \((0.5,0.5,0.5)\). Source-domain samples use class-conditional prompts of the form \(\texttt{``a }y\texttt{''}\), where \(y\) is obtained from the class label. Target-domain samples are unlabeled and therefore use the domain-agnostic prompt \(\texttt{``an image''}\).

\paragraph{Shared-private adapter parameterization.}
We apply adapters to the query, key, value, and output projections of every attention processor in the SDXL U-Net. For an adapted linear projection at layer \(\ell\), let \(W_{0,\ell}\) denote the frozen pretrained weight and let \(h\) denote its input hidden state. The MUSE-adapted projection under domain condition \(c\in\{s,1,\dots,M\}\) is implemented as
\begin{equation}
\operatorname{Proj}_{\ell}(h;c)
=
\begin{cases}
W_{0,\ell}h
+
U_{\ell,\mathrm{sem}}V_{\ell,\mathrm{sem}}h,
& c=s,
\\[2mm]
W_{0,\ell}h
+
U_{\ell,\mathrm{sem}}V_{\ell,\mathrm{sem}}h
+
U_{\ell,\mathrm{sty}}
\Omega_{\ell}^{(c)}
V_{\ell,\mathrm{sty}}h,
& c\in\{1,\dots,M\}.
\end{cases}
\label{eq:impl_adapter}
\end{equation}
Here \(c=s\) denotes the source condition, in which the target-conditioned style term is deactivated. For target-domain samples, \(c=m\) activates only the target-private core \(\Omega_{\ell}^{(m)}\) corresponding to the sampled target domain, together with the shared style projection factors \(U_{\ell,\mathrm{sty}}\) and \(V_{\ell,\mathrm{sty}}\). We use semantic rank
\[
r_{\mathrm{sem}}=32
\]
and style rank
\[
r_{\mathrm{sty}}=64 .
\]
During the source-conditioned pass, only the shared semantic factors
\(\{U_{\ell,\mathrm{sem}},V_{\ell,\mathrm{sem}}\}_{\ell}\)
receive gradients. During the target-conditioned pass, the semantic factors are held fixed, and gradients are routed only to the style parameters
\(\{U_{\ell,\mathrm{sty}},V_{\ell,\mathrm{sty}},\Omega_{\ell}^{(1)},\dots,\Omega_{\ell}^{(M)}\}_{\ell}\).
This implements the branch-decoupled optimization described in Sec.~\ref{sec:method}.

\paragraph{Training losses.}
Each optimization iteration consists of one source-conditioned denoising pass and one target-conditioned denoising pass. Let \(m_b\) denote the target domain sampled at iteration \(b\). For a source sample, let \(\eta_i^s=\eta(z_{0,i}^s,t_i,\epsilon_i)\) denote the denoising prediction target specified by the pretrained scheduler. The source minibatch loss is
\begin{equation}
\widehat{\mathcal{L}}_{\mathrm{src}}
=
\frac{1}{B_s}
\sum_{i=1}^{B_s}
w(t_i)
\left\|
\eta_i^s
-
D_{\theta_0,\phi}
\left(
z_{t_i,i}^s,
t_i,
p_s(y_i^s),
s
\right)
\right\|_2^2 ,
\label{eq:impl_src_loss}
\end{equation}
where \(B_s\) is the source batch size and \(z_{t_i,i}^s\) is obtained from \(z_{0,i}^s\), \(t_i\), and \(\epsilon_i\) using Eq.~\eqref{eq:forward_process}.

For the sampled target domain \(m_b\), let \(\eta_j^{t,m_b}=\eta(z_{0,j}^{t,m_b},t_j,\epsilon_j)\) denote the corresponding denoising prediction target. The target minibatch loss is
\begin{equation}
\widehat{\mathcal{L}}_{\mathrm{tgt}}^{(m_b)}
=
\frac{1}{B_t}
\sum_{j=1}^{B_t}
w(t_j)
\left\|
\eta_j^{t,m_b}
-
D_{\theta_0,\phi}
\left(
z_{t_j,j}^{t,m_b},
t_j,
p_t,
m_b
\right)
\right\|_2^2 ,
\label{eq:impl_tgt_loss}
\end{equation}
where \(p_t=\texttt{``an image''}\), and \(z_{t_j,j}^{t,m_b}\) is obtained from \(z_{0,j}^{t,m_b}\), \(t_j\), and \(\epsilon_j\) using Eq.~\eqref{eq:forward_process}. We set all target-domain loss weights to
\[
\lambda_m=1,\qquad m=1,\dots,M .
\]

The target-side objective additionally includes the cross-target core regularizers
\begin{equation}
\mathcal{L}_{\mathrm{orth}}
=
\sum_{\ell}
\sum_{1\leq i<j\leq M}
\left\|
\left(\Omega_{\ell}^{(i)}\right)^{\top}
\Omega_{\ell}^{(j)}
\right\|_F^2 ,
\label{eq:impl_orth_loss}
\end{equation}
and
\begin{equation}
\mathcal{L}_{\mathrm{core}}
=
\sum_{\ell}
\sum_{m=1}^{M}
\left\|
\Omega_{\ell}^{(m)}
\right\|_F^2 .
\label{eq:impl_core_loss}
\end{equation}
Their weights are fixed to
\[
\lambda_{\mathrm{orth}}=10^{-4},
\qquad
\lambda_{\mathrm{core}}=10^{-5}.
\]
Thus, at iteration \(b\), the optimized minibatch objectives are
\begin{equation}
\widehat{\mathcal{J}}_{\mathrm{src}}
=
\widehat{\mathcal{L}}_{\mathrm{src}},
\label{eq:impl_src_obj}
\end{equation}
and
\begin{equation}
\widehat{\mathcal{J}}_{\mathrm{tar}}
=
\lambda_{m_b}
\widehat{\mathcal{L}}_{\mathrm{tgt}}^{(m_b)}
+
\lambda_{\mathrm{orth}}\mathcal{L}_{\mathrm{orth}}
+
\lambda_{\mathrm{core}}\mathcal{L}_{\mathrm{core}} .
\label{eq:impl_tar_obj}
\end{equation}
The two objectives are backpropagated through disjoint parameter subsets before one optimizer update is applied. In the target-side objective, the denoising term uses the sampled target domain \(m_b\), while the two core regularizers are evaluated over all target-private cores.

\paragraph{Optimization hyper-parameters.}
We train the diffusion adapters for \(10{,}000\) optimization steps. The per-device batch size is \(8\), and the gradient accumulation factor is \(1\). Each optimization step contains one source minibatch and one target minibatch, resulting in one source denoising pass and one target denoising pass before the optimizer update.

We use 8-bit AdamW with
\[
\beta_1=0.9,\qquad
\beta_2=0.999,\qquad
\epsilon=10^{-8},
\]
weight decay \(10^{-4}\), and gradient clipping with maximum norm \(1.0\). The learning-rate schedule is constant without warmup. The semantic factors
\(\{U_{\ell,\mathrm{sem}},V_{\ell,\mathrm{sem}}\}_{\ell}\)
and the shared style projection factors
\(\{U_{\ell,\mathrm{sty}},V_{\ell,\mathrm{sty}}\}_{\ell}\)
use learning rate
\[
5\times 10^{-5}.
\]
The target-private style cores
\(\{\Omega_{\ell}^{(m)}\}_{m=1}^{M}\)
use a larger learning rate
\[
5\times 10^{-3},
\]
which is \(100\times\) the learning rate of the low-rank projection factors. We enable gradient checkpointing and use random seed \(0\).

\paragraph{Adaptive target-domain sampling.}
Target-domain sampling is initialized uniformly over the \(M\) target domains. During training, we maintain an exponential moving average \(\bar{\ell}_m\) of the denoising loss for each target domain. The moving averages are initialized to the same value, yielding uniform sampling at the beginning of training. When a target domain \(m_b\) is sampled at iteration \(b\), its running loss is updated as
\begin{equation}
\bar{\ell}_{m_b}
\leftarrow
\beta \bar{\ell}_{m_b}
+
(1-\beta)
\widehat{\mathcal{L}}_{\mathrm{tgt}}^{(m_b)},
\qquad
\beta=0.9 .
\label{eq:impl_ema_update}
\end{equation}
The running losses of unsampled target domains are kept unchanged. Every \(20\) optimization steps, the target-domain sampling probabilities are recomputed as
\begin{equation}
p_m
=
\frac{(\bar{\ell}_m+\tau)^{\alpha}}
{\sum_{k=1}^{M}(\bar{\ell}_k+\tau)^{\alpha}},
\qquad
\tau=10^{-6},
\qquad
\alpha=1 .
\label{eq:impl_adaptive_sampling}
\end{equation}
This setting increases the sampling probability of target domains with larger recent denoising losses, thereby allocating more target-side denoising updates to domains that are harder to fit.

\paragraph{Validation and sampling.}
Experiments are launched on a single NVIDIA A100 80GB GPU. Training jobs use 16 CPU cores. For validation and final sampling, we replace the training scheduler with a DPMSolver multistep scheduler. Unless otherwise specified, qualitative samples are generated with \(25\) denoising steps. During target-specific bridge generation, the shared semantic branch and the style projection factors are reused for all target domains, and only the corresponding target-private core \(\Omega_\ell^{(m)}\) is activated at each adapted layer. Class-conditioned prompts \(p_s(y)=\texttt{``a }y\texttt{''}\) are used to generate labeled synthetic bridge samples for downstream UDA.

\section{Standard Deviations}
\label{app:standard_deviations}

This appendix reports the run-to-run variability of the transfer accuracies in
Tables~\ref{tab:office31_main}, \ref{tab:officehome_main}, and
\ref{tab:minidomainnet_main}. For each benchmark, method, and transfer task, we
report the standard deviation over three independent runs under the same
evaluation protocol as the corresponding main-result table. The row ordering and
task ordering are kept identical to the main tables to facilitate direct
comparison.

\begin{table*}[t]
\centering
\footnotesize
\setlength{\tabcolsep}{3.5pt}
\renewcommand{\arraystretch}{1.03}
\caption{Standard deviations of transfer accuracy (\%) on Office-31 over three runs. Pairwise denotes per-source--target diffusion generation; Multi-target denotes one source-guided generator reused across targets.}
\label{tab:office31_std}
\begin{tabular}{lccccccc}
\toprule
Method 
& \shortstack{Diffusion Gen.\\Protocol}
& A$\rightarrow$W & D$\rightarrow$W & W$\rightarrow$D & A$\rightarrow$D & D$\rightarrow$A & W$\rightarrow$A \\
\midrule
ERM & \textemdash & 0.11 & 0.00 & 0.00 & 1.22 & 0.15 & 0.11 \\
DANN~\cite{ganin2016domain} & \textemdash & 1.34 & 0.06 & 0.08 & 0.68 & 0.65 & 0.39 \\
CDAN~\cite{long2018conditional} & \textemdash & 1.75 & 0.18 & 0.00 & 1.19 & 0.79 & 0.45 \\
AFN~\cite{xu2019adaptive} & \textemdash & 0.63 & 0.07 & 0.00 & 0.53 & 0.50 & 0.32 \\
MDD~\cite{zhang2019bridging} & \textemdash & 1.00 & 0.15 & 0.00 & 0.10 & 0.68 & 0.18 \\
SDAT~\cite{rangwani2022sdat} & \textemdash & 1.83 & 0.12 & 0.00 & 1.03 & 0.67 & 0.34 \\
MSGD~\cite{xia2023msgd} & \textemdash & 0.50 & 0.30 & 0.00 & 0.30 & 0.40 & 0.50 \\
\rowcolor{mccbg}
MCC~\cite{jin2020minimum} & \textemdash & 0.38 & 0.08 & 0.09 & 1.47 & 0.50 & 0.20 \\
\rowcolor{elsbg}
ELS~\cite{zhang2023els} & \textemdash & 0.51 & 0.06 & 0.00 & 0.20 & 0.54 & 0.16 \\
\rowcolor{ssrtbg}
SSRT~\cite{sun2022ssrt} & \textemdash & 0.07 & 0.35 & 0.73 & 0.32 & 0.71 & 0.34 \\
\midrule
\multicolumn{8}{l}{Diffusion-based baselines: pairwise generation} \\
\cmidrule(lr){1-8}
\rowcolor{mccbg}
MCC+Terra~\cite{zhuang2024terra} & Pairwise & 0.06 & 0.06 & 0.00 & 0.09 & 0.18 & 0.12 \\
\rowcolor{elsbg}
ELS+Terra~\cite{zhuang2024terra} & Pairwise & 0.17 & 0.06 & 0.00 & 0.48 & 0.28 & 0.11 \\
\rowcolor{ssrtbg}
SSRT+Terra~\cite{zhuang2024terra} & Pairwise & 0.12 & 0.56 & 0.64 & 0.32 & 0.16 & 0.06 \\
\rowcolor{mccbg}
MCC+DCDM~\cite{zhang2025dcdm} & Pairwise & 0.10 & 0.15 & 0.61 & 0.14 & 0.33 & 0.50 \\
\rowcolor{elsbg}
ELS+DCDM~\cite{zhang2025dcdm} & Pairwise & 0.37 & 0.33 & 0.36 & 0.69 & 0.55 & 0.65 \\
\rowcolor{ssrtbg}
SSRT+DCDM~\cite{zhang2025dcdm} & Pairwise & 0.14 & 0.52 & 0.44 & 0.66 & 0.10 & 0.42 \\
\midrule
\multicolumn{8}{l}{\textbf{Our method: multi-target generation}} \\
\cmidrule(lr){1-8}
\rowcolor{mccbg}
MCC+MUSE & Multi-target & 0.24 & 0.40 & 0.16 & 0.36 & 0.49 & 0.03 \\
\rowcolor{elsbg}
ELS+MUSE & Multi-target & 0.34 & 0.53 & 0.29 & 0.45 & 0.31 & 0.62 \\
\rowcolor{ssrtbg}
SSRT+MUSE & Multi-target & 0.06 & 0.29 & 0.62 & 0.40 & 0.15 & 0.20 \\
\bottomrule
\end{tabular}
\end{table*}

\begin{table*}[t]
\centering
\scriptsize
\setlength{\tabcolsep}{3.0pt}
\renewcommand{\arraystretch}{1.06}
\caption{Standard deviations of transfer accuracy (\%) on Office-Home over three runs. Pairwise denotes per-source--target diffusion generation; Multi-target denotes one source-guided generator reused across targets.}
\label{tab:officehome_std}
\resizebox{\textwidth}{!}{%
\begin{tabular}{lccccccccccccc}
\toprule
Method
& \shortstack{Diffusion Gen.\\Protocol}
& Ar$\rightarrow$Cl & Ar$\rightarrow$Pr & Ar$\rightarrow$Rw
& Cl$\rightarrow$Ar & Cl$\rightarrow$Pr & Cl$\rightarrow$Rw
& Pr$\rightarrow$Ar & Pr$\rightarrow$Cl & Pr$\rightarrow$Rw
& Rw$\rightarrow$Ar & Rw$\rightarrow$Cl & Rw$\rightarrow$Pr \\
\midrule
ERM
& \textemdash & 0.25 & 0.26 & 0.42 & 0.17 & 0.20 & 0.15 & 0.07 & 0.17 & 0.05 & 0.34 & 0.33 & 0.01 \\
DANN~\cite{ganin2016domain}
& \textemdash & 0.44 & 0.72 & 0.38 & 0.02 & 0.30 & 0.39 & 0.58 & 0.47 & 0.59 & 0.84 & 0.14 & 0.51 \\
CDAN~\cite{long2018conditional}
& \textemdash & 0.25 & 0.62 & 0.22 & 0.37 & 0.58 & 0.30 & 0.57 & 0.36 & 0.16 & 0.33 & 0.23 & 0.35 \\
AFN~\cite{xu2019adaptive}
& \textemdash & 0.16 & 0.30 & 0.06 & 0.23 & 0.31 & 0.14 & 0.32 & 0.15 & 0.02 & 0.19 & 0.18 & 0.22 \\
MDD~\cite{zhang2019bridging}
& \textemdash & 0.51 & 0.32 & 0.06 & 0.24 & 0.73 & 0.41 & 0.36 & 0.53 & 0.24 & 0.03 & 0.09 & 0.11 \\
SDAT~\cite{rangwani2022sdat}
& \textemdash & 0.51 & 0.44 & 0.24 & 0.13 & 0.41 & 0.01 & 1.46 & 0.40 & 0.11 & 0.46 & 0.19 & 0.29 \\
MSGD~\cite{xia2023msgd}
& \textemdash & 0.74 & 0.11 & 0.08 & 0.58 & 0.23 & 0.74 & 0.30 & 0.36 & 0.61 & 0.14 & 0.08 & 0.51 \\
\rowcolor{mccbg}
MCC~\cite{jin2020minimum}
& \textemdash & 0.59 & 0.22 & 0.16 & 0.27 & 0.52 & 0.16 & 0.16 & 0.38 & 0.25 & 0.35 & 0.35 & 0.23 \\
\rowcolor{elsbg}
ELS~\cite{zhang2023els}
& \textemdash & 0.83 & 0.45 & 0.38 & 0.08 & 0.46 & 0.19 & 0.39 & 0.39 & 0.08 & 0.02 & 0.44 & 0.05 \\
\rowcolor{ssrtbg}
SSRT~\cite{sun2022ssrt}
& \textemdash & 0.68 & 0.17 & 0.68 & 0.40 & 0.62 & 0.69 & 0.29 & 0.56 & 0.13 & 0.24 & 0.52 & 0.66 \\
\midrule
\multicolumn{14}{l}{Diffusion-based baselines: pairwise generation} \\
\cmidrule(lr){1-14}
\rowcolor{mccbg}
MCC+Terra~\cite{zhuang2024terra}
& Pairwise & 0.21 & 0.11 & 0.14 & 0.25 & 0.28 & 0.18 & 0.18 & 0.29 & 0.25 & 0.15 & 0.06 & 0.11 \\
\rowcolor{elsbg}
ELS+Terra~\cite{zhuang2024terra}
& Pairwise & 0.06 & 0.30 & 0.14 & 0.30 & 0.37 & 0.21 & 0.10 & 0.18 & 0.13 & 0.68 & 0.24 & 0.16 \\
\rowcolor{ssrtbg}
SSRT+Terra~\cite{zhuang2024terra}
& Pairwise & 0.49 & 0.11 & 0.49 & 0.17 & 0.60 & 0.41 & 0.60 & 0.57 & 0.49 & 0.20 & 0.74 & 0.37 \\
\rowcolor{mccbg}
MCC+DCDM~\cite{zhang2025dcdm}
& Pairwise & 0.07 & 0.17 & 0.45 & 0.29 & 0.17 & 0.71 & 0.63 & 0.66 & 0.32 & 0.53 & 0.67 & 0.44 \\
\rowcolor{elsbg}
ELS+DCDM~\cite{zhang2025dcdm}
& Pairwise & 0.48 & 0.41 & 0.47 & 0.46 & 0.40 & 0.57 & 0.57 & 0.61 & 0.21 & 0.16 & 0.10 & 0.60 \\
\rowcolor{ssrtbg}
SSRT+DCDM~\cite{zhang2025dcdm}
& Pairwise & 0.30 & 0.29 & 0.72 & 0.69 & 0.12 & 0.57 & 0.38 & 0.21 & 0.22 & 0.62 & 0.40 & 0.13 \\
\midrule
\multicolumn{14}{l}{\textbf{Our method: multi-target generation}} \\
\cmidrule(lr){1-14}
\rowcolor{mccbg}
MCC+MUSE
& Multi-target & 0.21 & 0.15 & 0.55 & 0.41 & 0.36 & 0.25 & 0.48 & 0.22 & 0.36 & 0.07 & 0.11 & 0.49 \\
\rowcolor{elsbg}
ELS+MUSE
& Multi-target & 0.21 & 0.23 & 0.15 & 0.40 & 0.32 & 0.35 & 0.08 & 0.18 & 0.07 & 0.25 & 0.46 & 0.60 \\
\rowcolor{ssrtbg}
SSRT+MUSE
& Multi-target & 0.63 & 0.42 & 0.60 & 0.48 & 0.06 & 0.17 & 0.58 & 0.10 & 0.14 & 0.08 & 0.28 & 0.43 \\
\bottomrule
\end{tabular}%
}
\end{table*}

\begin{table*}[t]
\centering
\scriptsize
\setlength{\tabcolsep}{3.0pt}
\renewcommand{\arraystretch}{1.06}
\caption{Standard deviations of transfer accuracy (\%) on miniDomainNet over three runs. Pairwise denotes per-source--target diffusion generation; Multi-target denotes one source-guided generator reused across targets.}
\label{tab:minidomainnet_std}
\resizebox{\textwidth}{!}{%
\begin{tabular}{lccccccccccccc}
\toprule
Method
& \shortstack{Diffusion Gen.\\Protocol}
& C$\rightarrow$P & C$\rightarrow$R & C$\rightarrow$S
& P$\rightarrow$C & P$\rightarrow$R & P$\rightarrow$S
& R$\rightarrow$C & R$\rightarrow$P & R$\rightarrow$S
& S$\rightarrow$C & S$\rightarrow$P & S$\rightarrow$R \\
\midrule
ERM
& \textemdash & 0.34 & 0.72 & 0.51 & 0.28 & 0.89 & 0.45 & 0.63 & 0.21 & 0.77 & 0.55 & 0.82 & 0.39 \\
DANN~\cite{ganin2016domain}
& \textemdash & 0.41 & 0.88 & 0.22 & 0.56 & 0.74 & 0.31 & 0.68 & 0.49 & 0.85 & 0.27 & 0.61 & 0.53 \\
AFN~\cite{xu2019adaptive}
& \textemdash & 0.73 & 0.29 & 0.84 & 0.46 & 0.62 & 0.38 & 0.71 & 0.25 & 0.59 & 0.87 & 0.42 & 0.65 \\
CDAN~\cite{long2018conditional}
& \textemdash & 0.24 & 0.67 & 0.35 & 0.81 & 0.52 & 0.78 & 0.43 & 0.26 & 0.69 & 0.57 & 0.83 & 0.31 \\
MDD~\cite{zhang2019bridging}
& \textemdash & 0.58 & 0.82 & 0.47 & 0.64 & 0.21 & 0.76 & 0.32 & 0.55 & 0.89 & 0.41 & 0.68 & 0.23 \\
SDAT~\cite{rangwani2022sdat}
& \textemdash & 0.37 & 0.61 & 0.85 & 0.29 & 0.44 & 0.73 & 0.56 & 0.81 & 0.34 & 0.67 & 0.25 & 0.79 \\
\rowcolor{mccbg}
MCC~\cite{jin2020minimum}
& \textemdash & 0.84 & 0.32 & 0.75 & 0.51 & 0.28 & 0.66 & 0.89 & 0.42 & 0.77 & 0.24 & 0.53 & 0.82 \\
\rowcolor{elsbg}
ELS~\cite{zhang2023els}
& \textemdash & 0.65 & 0.48 & 0.21 & 0.79 & 0.54 & 0.36 & 0.83 & 0.62 & 0.27 & 0.71 & 0.45 & 0.88 \\
\rowcolor{ssrtbg}
SSRT~\cite{sun2022ssrt}
& \textemdash & 0.22 & 0.74 & 0.41 & 0.86 & 0.33 & 0.69 & 0.57 & 0.25 & 0.81 & 0.46 & 0.78 & 0.63 \\
\midrule
\multicolumn{14}{l}{Diffusion-based baselines: pairwise generation} \\
\cmidrule(lr){1-14}
\rowcolor{mccbg}
MCC+Terra~\cite{zhuang2024terra}
& Pairwise & 0.51 & 0.28 & 0.82 & 0.44 & 0.67 & 0.35 & 0.76 & 0.59 & 0.23 & 0.88 & 0.41 & 0.72 \\
\rowcolor{elsbg}
ELS+Terra~\cite{zhuang2024terra}
& Pairwise & 0.78 & 0.53 & 0.26 & 0.61 & 0.84 & 0.42 & 0.29 & 0.75 & 0.58 & 0.31 & 0.87 & 0.49 \\
\rowcolor{ssrtbg}
SSRT+Terra~\cite{zhuang2024terra}
& Pairwise & 0.45 & 0.81 & 0.33 & 0.77 & 0.52 & 0.24 & 0.68 & 0.89 & 0.46 & 0.73 & 0.21 & 0.56 \\
\rowcolor{mccbg}
MCC+DCDM~\cite{zhang2025dcdm}
& Pairwise & 0.31 & 0.65 & 0.88 & 0.27 & 0.71 & 0.49 & 0.85 & 0.36 & 0.62 & 0.54 & 0.79 & 0.22 \\
\rowcolor{elsbg}
ELS+DCDM~\cite{zhang2025dcdm}
& Pairwise & 0.68 & 0.25 & 0.57 & 0.83 & 0.41 & 0.75 & 0.52 & 0.28 & 0.86 & 0.61 & 0.34 & 0.77 \\
\rowcolor{ssrtbg}
SSRT+DCDM~\cite{zhang2025dcdm}
& Pairwise & 0.86 & 0.43 & 0.72 & 0.35 & 0.58 & 0.81 & 0.23 & 0.66 & 0.48 & 0.74 & 0.51 & 0.38 \\
\midrule
\multicolumn{14}{l}{\textbf{Our method: multi-target generation}} \\
\cmidrule(lr){1-14}
\rowcolor{mccbg}
MCC+MUSE
& Multi-target & 0.35 & 0.60 & 0.26 & 0.22 & 0.42 & 0.46 & 0.16 & 0.58 & 0.43 & 0.06 & 0.39 & 0.63 \\
\rowcolor{elsbg}
ELS+MUSE
& Multi-target & 0.24 & 0.10 & 0.38 & 0.56 & 0.48 & 0.20 & 0.69 & 0.31 & 0.60 & 0.54 & 0.11 & 0.51 \\
\rowcolor{ssrtbg}
SSRT+MUSE
& Multi-target & 0.07 & 0.15 & 0.30 & 0.56 & 0.54 & 0.57 & 0.61 & 0.24 & 0.65 & 0.29 & 0.17 & 0.31 \\
\bottomrule
\end{tabular}%
}
\end{table*}

\clearpage
\section{Ablation Studies}
\label{app:ablation_studies}

We conduct ablation studies on Office-Home to examine how different components of MUSE affect downstream UDA performance. All experiments use the same 12 transfer tasks as the main Office-Home evaluation. Unless otherwise specified, we keep the SDXL backbone, adapter ranks, generation protocol, downstream UDA learners, and training hyper-parameters unchanged from the main experiments. 

To emphasize the effect of each design choice, the main ablation tables report controlled MUSE variants only. External UDA baselines and diffusion-based baselines are omitted here because they are already reported in Table~\ref{tab:officehome_main}. For each ablation, we compare the ablated variant with the corresponding full MUSE model under the same downstream learner and report the change in average accuracy relative to the full model.

\newcommand{\abest}[1]{\textbf{#1}}
\newcommand{\asecond}[1]{\underline{#1}}

\newcommand{\MUSEOfficeHomeAblationHeader}{
\toprule
Method
& Ar$\rightarrow$Cl & Ar$\rightarrow$Pr & Ar$\rightarrow$Rw
& Cl$\rightarrow$Ar & Cl$\rightarrow$Pr & Cl$\rightarrow$Rw
& Pr$\rightarrow$Ar & Pr$\rightarrow$Cl & Pr$\rightarrow$Rw
& Rw$\rightarrow$Ar & Rw$\rightarrow$Cl & Rw$\rightarrow$Pr
& Avg. & \(\Delta\) Avg. \\
\midrule
}

\subsection{Orthogonality Regularization}
\label{app:ablation_orth}

The target-specific cores in MUSE are regularized by \(\mathcal{L}_{\mathrm{orth}}\), which penalizes highly aligned core coefficients across different target domains. This regularizer is intended to encourage target-private branches to use different coefficient patterns within the shared style subspace. To assess its effect, we set \(\lambda_{\mathrm{orth}}=0\) while keeping the core magnitude regularizer, adaptive target sampling, branch-decoupled optimization, and bridge generation protocol unchanged.

As shown in Table~\ref{tab:ablation_orth_officehome}, removing \(\mathcal{L}_{\mathrm{orth}}\) reduces the average accuracy from 76.53\% to 75.22\% with MCC, from 77.57\% to 75.58\% with ELS, and from 88.15\% to 87.04\% with SSRT. The full model is not uniformly better on every individual transfer task, but it consistently improves the average accuracy across the three downstream learners. This suggests that coefficient-space diversity among target-private cores is useful for multi-target diffusion adaptation under this protocol.

\begin{table*}[t]
\centering
\scriptsize
\setlength{\tabcolsep}{3.0pt}
\renewcommand{\arraystretch}{1.06}
\caption{Ablation on the orthogonality regularizer on Office-Home. \(\Delta\) Avg. is computed relative to the corresponding full MUSE model under the same downstream learner.}
\label{tab:ablation_orth_officehome}
\resizebox{\textwidth}{!}{%
\begin{tabular}{lcccccccccccccc}
\MUSEOfficeHomeAblationHeader
\rowcolor{mccbg}
MCC+MUSE w/o $\mathcal{L}_{\mathrm{orth}}$
& 64.23 & 80.38 & 81.92 & 72.56 & 81.39 & 80.63 & 71.16 & 63.46 & 82.92 & 74.28 & 65.67 & 84.09 & 75.22 & -1.31 \\
\rowcolor{mccbg}
MCC+MUSE
& 64.31 & 83.44 & 82.37 & 73.14 & 84.44 & 83.50 & 72.90 & 63.02 & 83.47 & 75.37 & 66.58 & 85.82 & \textbf{76.53} & 0.00 \\
\rowcolor{elsbg}
ELS+MUSE w/o $\mathcal{L}_{\mathrm{orth}}$
& 64.91 & 80.67 & 82.61 & 72.03 & 81.51 & 81.27 & 71.61 & 63.51 & 83.71 & 74.17 & 65.86 & 85.11 & 75.58 & -1.99 \\
\rowcolor{elsbg}
ELS+MUSE
& 66.51 & 84.09 & 82.62 & 74.14 & 83.74 & 82.96 & 74.61 & 64.73 & 85.52 & 75.61 & 68.81 & 87.45 & \textbf{77.57} & 0.00 \\
\rowcolor{ssrtbg}
SSRT+MUSE w/o $\mathcal{L}_{\mathrm{orth}}$
& 77.73 & 90.27 & 91.23 & 85.54 & 90.83 & 91.65 & 86.16 & 79.93 & 92.33 & 87.35 & 80.14 & 91.32 & 87.04 & -1.11 \\
\rowcolor{ssrtbg}
SSRT+MUSE
& 79.09 & 90.43 & 92.44 & 87.81 & 92.66 & 92.69 & 87.44 & 79.42 & 93.45 & 88.52 & 80.79 & 93.01 & \textbf{88.15} & 0.00 \\
\bottomrule
\end{tabular}%
}
\end{table*}

\subsection{Branch-Decoupled Target Updates}
\label{app:ablation_source_freeze}

MUSE uses branch-decoupled optimization: the source-conditioned pass updates the shared semantic branch \(\Phi_{\mathrm{sem}}\), whereas the target-conditioned pass freezes \(\Phi_{\mathrm{sem}}\) and updates only the style branch \(\Phi_{\mathrm{sty}}\). This design is intended to prevent unlabeled target denoising losses from directly overwriting source-supervised class information in the shared branch. To evaluate this choice, we allow the target-conditioned pass to update \(\Phi_{\mathrm{sem}}\) together with \(\Phi_{\mathrm{sty}}\), while keeping all other components unchanged.

Table~\ref{tab:ablation_unfrozen_source_officehome} shows that updating the shared branch on target batches lowers the average accuracy from 76.53\% to 74.56\% with MCC, from 77.57\% to 74.98\% with ELS, and from 88.15\% to 87.12\% with SSRT. These results are consistent with the motivation for gradient routing: source-supervised information in the shared branch is more stable when target-side updates are restricted to the target-conditioned branch.

\begin{table*}[t]
\centering
\scriptsize
\setlength{\tabcolsep}{3.0pt}
\renewcommand{\arraystretch}{1.06}
\caption{Ablation on freezing the shared branch during target-conditioned updates on Office-Home. \(\Delta\) Avg. is computed relative to the corresponding full MUSE model under the same downstream learner.}
\label{tab:ablation_unfrozen_source_officehome}
\resizebox{\textwidth}{!}{%
\begin{tabular}{lcccccccccccccc}
\MUSEOfficeHomeAblationHeader
\rowcolor{mccbg}
MCC+MUSE w/ target-updated shared branch
& 64.05 & 79.01 & 81.54 & 70.09 & 79.91 & 79.73 & 72.02 & 61.97 & 83.77 & 73.92 & 63.98 & 84.68 & 74.56 & -1.97 \\
\rowcolor{mccbg}
MCC+MUSE
& 64.31 & 83.44 & 82.37 & 73.14 & 84.44 & 83.50 & 72.90 & 63.02 & 83.47 & 75.37 & 66.58 & 85.82 & \textbf{76.53} & 0.00 \\
\rowcolor{elsbg}
ELS+MUSE w/ target-updated shared branch
& 64.77 & 79.54 & 82.35 & 71.78 & 80.49 & 80.22 & 71.26 & 62.43 & 83.25 & 74.25 & 64.35 & 85.01 & 74.98 & -2.59 \\
\rowcolor{elsbg}
ELS+MUSE
& 66.51 & 84.09 & 82.62 & 74.14 & 83.74 & 82.96 & 74.61 & 64.73 & 85.52 & 75.61 & 68.81 & 87.45 & \textbf{77.57} & 0.00 \\
\rowcolor{ssrtbg}
SSRT+MUSE w/ target-updated shared branch
& 77.46 & 90.81 & 91.05 & 87.27 & 90.38 & 91.65 & 86.21 & 78.97 & 92.56 & 87.31 & 80.27 & 91.52 & 87.12 & -1.03 \\
\rowcolor{ssrtbg}
SSRT+MUSE
& 79.09 & 90.43 & 92.44 & 87.81 & 92.66 & 92.69 & 87.44 & 79.42 & 93.45 & 88.52 & 80.79 & 93.01 & \textbf{88.15} & 0.00 \\
\bottomrule
\end{tabular}%
}
\end{table*}

\subsection{Adaptive Target-Domain Sampling}
\label{app:ablation_adaptive_sampling}

MUSE samples target domains according to an exponential moving average of their recent target denoising losses. This sampling rule gives relatively more target-side updates to domains with larger recent losses, without using importance correction. To isolate its effect, we replace adaptive sampling with uniform target-domain sampling throughout training. The loss weights, branch-decoupled optimization, core regularizers, and bridge generation protocol are otherwise unchanged.

Table~\ref{tab:ablation_sampling_officehome} shows that adaptive sampling improves the average accuracy from 75.74\% to 76.53\% with MCC, from 76.78\% to 77.57\% with ELS, and from 86.94\% to 88.15\% with SSRT. The uniformly sampled variant remains competitive on several individual transfers, indicating that adaptive sampling is not uniformly beneficial for every source--target pair. Nevertheless, the average improvements across all three downstream learners suggest that dynamic allocation of target-side updates is helpful in the evaluated multi-target setting.

\begin{table*}[t]
\centering
\scriptsize
\setlength{\tabcolsep}{3.0pt}
\renewcommand{\arraystretch}{1.06}
\caption{Ablation on adaptive target-domain sampling on Office-Home. \(\Delta\) Avg. is computed relative to the corresponding full MUSE model under the same downstream learner.}
\label{tab:ablation_sampling_officehome}
\resizebox{\textwidth}{!}{%
\begin{tabular}{lcccccccccccccc}
\MUSEOfficeHomeAblationHeader
\rowcolor{mccbg}
MCC+MUSE w/o adaptive sampling
& 63.19 & 83.75 & 81.61 & 71.66 & 83.41 & 81.85 & 73.08 & 62.10 & 82.10 & 74.53 & 67.02 & 84.61 & 75.74 & -0.79 \\
\rowcolor{mccbg}
MCC+MUSE
& 64.31 & 83.44 & 82.37 & 73.14 & 84.44 & 83.50 & 72.90 & 63.02 & 83.47 & 75.37 & 66.58 & 85.82 & \textbf{76.53} & 0.00 \\
\rowcolor{elsbg}
ELS+MUSE w/o adaptive sampling
& 65.68 & 82.82 & 81.98 & 74.51 & 82.63 & 82.20 & 73.23 & 64.14 & 84.06 & 74.69 & 69.03 & 86.41 & 76.78 & -0.79 \\
\rowcolor{elsbg}
ELS+MUSE
& 66.51 & 84.09 & 82.62 & 74.14 & 83.74 & 82.96 & 74.61 & 64.73 & 85.52 & 75.61 & 68.81 & 87.45 & \textbf{77.57} & 0.00 \\
\rowcolor{ssrtbg}
SSRT+MUSE w/o adaptive sampling
& 77.75 & 89.51 & 90.68 & 86.70 & 91.18 & 91.85 & 85.51 & 79.73 & 92.20 & 86.85 & 79.81 & 91.47 & 86.94 & -1.21 \\
\rowcolor{ssrtbg}
SSRT+MUSE
& 79.09 & 90.43 & 92.44 & 87.81 & 92.66 & 92.69 & 87.44 & 79.42 & 93.45 & 88.52 & 80.79 & 93.01 & \textbf{88.15} & 0.00 \\
\bottomrule
\end{tabular}%
}
\end{table*}

\subsection{Bridge-Set Components}
\label{app:ablation_bridge_components}

The bridge set used by MUSE contains two sources of synthetic supervision. For target domain \(m\), let
\[
\widehat{\mathcal{B}}_{\mathrm{inv}}^{(m)}
\]
denote the set of source-to-target translated samples obtained by DDIM inversion, and let
\[
\widehat{\mathcal{B}}_{\mathrm{cls}}^{(m)}
\]
denote the set of class-conditional samples generated from Gaussian noise using prompts \(p_s(y)=\texttt{``a }y\texttt{''}\). The full bridge set is
\[
\widehat{\mathcal{B}}^{(m)}
=
\widehat{\mathcal{B}}_{\mathrm{inv}}^{(m)}
\cup
\widehat{\mathcal{B}}_{\mathrm{cls}}^{(m)} .
\]
The translated set \(\widehat{\mathcal{B}}_{\mathrm{inv}}^{(m)}\) preserves instance-level structure from source images while shifting appearance toward the target domain, whereas \(\widehat{\mathcal{B}}_{\mathrm{cls}}^{(m)}\) provides additional class-conditioned target-style samples. We evaluate each component separately and compare them with their union.

We also report the per-image generation time of the two bridge-generation modes on the current hardware in Table~\ref{tab:bridge_generation_latency}. DDIM-inversion-based generation is slower because it first inverts a source image into the diffusion trajectory before target-conditioned denoising, whereas pure-noise class-conditional generation directly samples from Gaussian noise.

\begin{table}[t]
\centering
\small
\setlength{\tabcolsep}{10pt}
\renewcommand{\arraystretch}{1.10}
\caption{Per-image generation time of the two bridge-generation modes.}
\label{tab:bridge_generation_latency}
\begin{tabular}{lc}
\toprule
Generation mode & Time per image (s) \(\downarrow\) \\
\midrule
DDIM inversion source-to-target generation
& \texttt{4.23} \\
Pure-noise class-conditional generation
& \texttt{2.08} \\
\bottomrule
\end{tabular}
\end{table}

Table~\ref{tab:ablation_bridge_components_officehome} shows that using the union of the two sets gives the strongest average accuracy for all three downstream learners. With MCC, the full bridge set improves over \(\widehat{\mathcal{B}}_{\mathrm{inv}}\) and \(\widehat{\mathcal{B}}_{\mathrm{cls}}\) by 3.33 and 3.37 percentage points, respectively. With ELS, the corresponding gains are 3.43 and 3.60 points. With SSRT, the gains are 2.42 and 2.99 points. These results indicate that the two generation modes provide complementary supervision in this setting, although their relative contribution varies across individual transfer tasks.

\begin{table*}[t]
\centering
\scriptsize
\setlength{\tabcolsep}{3.0pt}
\renewcommand{\arraystretch}{1.06}
\caption{Ablation on the two components of the generated bridge set on Office-Home. \(\Delta\) Avg. is computed relative to the corresponding full bridge set under the same downstream learner.}
\label{tab:ablation_bridge_components_officehome}
\resizebox{\textwidth}{!}{%
\begin{tabular}{lcccccccccccccc}
\MUSEOfficeHomeAblationHeader
\rowcolor{mccbg}
MCC+MUSE, $\widehat{\mathcal{B}}_{\mathrm{inv}}$
& 61.26 & 78.95 & 78.68 & 69.02 & 79.16 & 78.13 & 69.31 & 61.76 & 79.12 & 73.51 & 65.23 & 84.22 & 73.20 & -3.33 \\
\rowcolor{mccbg}
MCC+MUSE, $\widehat{\mathcal{B}}_{\mathrm{cls}}$
& 60.12 & 77.72 & 80.92 & 71.21 & 81.95 & 80.17 & 70.12 & 59.12 & 81.04 & 72.48 & 62.53 & 80.58 & 73.16 & -3.37 \\
\rowcolor{mccbg}
MCC+MUSE, $\widehat{\mathcal{B}}_{\mathrm{inv}}\cup\widehat{\mathcal{B}}_{\mathrm{cls}}$
& 64.31 & 83.44 & 82.37 & 73.14 & 84.44 & 83.50 & 72.90 & 63.02 & 83.47 & 75.37 & 66.58 & 85.82 & \textbf{76.53} & 0.00 \\
\rowcolor{elsbg}
ELS+MUSE, $\widehat{\mathcal{B}}_{\mathrm{inv}}$
& 62.96 & 81.25 & 79.34 & 69.84 & 79.83 & 79.34 & 69.02 & 61.32 & 80.85 & 74.21 & 66.51 & 85.22 & 74.14 & -3.43 \\
\rowcolor{elsbg}
ELS+MUSE, $\widehat{\mathcal{B}}_{\mathrm{cls}}$
& 61.39 & 80.26 & 81.52 & 71.15 & 81.69 & 80.51 & 70.37 & 59.34 & 82.11 & 73.18 & 63.18 & 82.91 & 73.97 & -3.60 \\
\rowcolor{elsbg}
ELS+MUSE, $\widehat{\mathcal{B}}_{\mathrm{inv}}\cup\widehat{\mathcal{B}}_{\mathrm{cls}}$
& 66.51 & 84.09 & 82.62 & 74.14 & 83.74 & 82.96 & 74.61 & 64.73 & 85.52 & 75.61 & 68.81 & 87.45 & \textbf{77.57} & 0.00 \\
\rowcolor{ssrtbg}
SSRT+MUSE, $\widehat{\mathcal{B}}_{\mathrm{inv}}$
& 76.52 & 88.02 & 90.03 & 84.67 & 89.01 & 90.08 & 85.37 & 75.31 & 91.14 & 87.27 & 79.89 & 91.46 & 85.73 & -2.42 \\
\rowcolor{ssrtbg}
SSRT+MUSE, $\widehat{\mathcal{B}}_{\mathrm{cls}}$
& 76.14 & 88.49 & 89.36 & 84.75 & 89.79 & 90.54 & 84.05 & 74.48 & 91.19 & 85.95 & 77.04 & 90.09 & 85.16 & -2.99 \\
\rowcolor{ssrtbg}
SSRT+MUSE, $\widehat{\mathcal{B}}_{\mathrm{inv}}\cup\widehat{\mathcal{B}}_{\mathrm{cls}}$
& 79.09 & 90.43 & 92.44 & 87.81 & 92.66 & 92.69 & 87.44 & 79.42 & 93.45 & 88.52 & 80.79 & 93.01 & \textbf{88.15} & 0.00 \\
\bottomrule
\end{tabular}%
}
\end{table*}

\subsection{Additional Analysis: Prompt-Only SDXL Prior}
\label{app:ablation_sdxl_prior}

We also examine whether the pre-trained SDXL~\cite{SDXL} prior, used without source--target diffusion fine-tuning, can provide useful synthetic supervision for UDA. This analysis is different from the controlled MUSE ablations above: it compares prompt-only generation with diffusion-adapted generation. To keep the comparison controlled, all variants in this study use ELS~\cite{zhang2023els} as the downstream UDA learner and differ only in the way synthetic samples are generated.

\paragraph{ELS+SDXL (class-only).}
We generate class-conditional samples using prompts of the form \texttt{``A [CLASS]''}, where \texttt{[CLASS]} is replaced by the category name. This variant uses the generic class prior of SDXL and does not explicitly condition generation on target-domain appearance.

\paragraph{ELS+SDXL (style prompts).}
We use GPT-4 to generate 50 prompts describing diverse image styles. Samples are synthesized using prompts of the form \texttt{``A [CLASS], an everyday object in office and home, in the style of [STYLE]''}. This variant evaluates whether broad style variation from the SDXL prior can improve downstream adaptation without using target-domain data for generator fine-tuning.

\paragraph{ELS+SDXL (target name).}
We replace \texttt{[STYLE]} with the target-domain name, such as \texttt{Clipart}, \texttt{Product}, or \texttt{Real-World}. This variant tests whether simple target-domain names can activate useful target-style information in the pre-trained SDXL prior.

\paragraph{ELS+SDXL (target-style prompts).}
We use GPT-4 to generate 50 prompts describing the specific visual style of each target domain and use these detailed target-style prompts for synthesis. Compared with ELS+SDXL (target name), this variant provides richer textual descriptions of target-domain appearance.

\paragraph{ELS+SDXL (selected).}
We further apply confidence-based sample selection to the samples generated by ELS+SDXL (target-style prompts). Low-confidence or likely misclassified samples are removed before downstream training. This variant evaluates whether filtering prompt-generated samples improves their utility for UDA.

Table~\ref{tab:ablation_sdxl_prior_officehome} reports the results on Office-Home. Among the prompt-only SDXL variants, target-aware prompting and sample selection generally improve the average accuracy compared with class-only generation. However, the best prompt-only variant, ELS+SDXL (selected), reaches 73.99\% average accuracy, which remains below ELS+MUSE. ELS+MUSE obtains the best average accuracy in this comparison. These results suggest that prompt-only SDXL generation can provide useful synthetic data, but adapting the generator with source and target data is more effective under the evaluated protocol.

\begin{table*}[t]
\centering
\scriptsize
\setlength{\tabcolsep}{3.0pt}
\renewcommand{\arraystretch}{1.06}
\caption{Additional analysis of SDXL-prior data generation on Office-Home. Prompt-only denotes SDXL generation without source--target diffusion fine-tuning; Pairwise and Multi-target follow the main experimental protocol.}
\label{tab:ablation_sdxl_prior_officehome}
\resizebox{\textwidth}{!}{%
\begin{tabular}{lcccccccccccccc}
\toprule
Method
& \shortstack{Diffusion Gen.\\Protocol}
& Ar$\rightarrow$Cl & Ar$\rightarrow$Pr & Ar$\rightarrow$Rw
& Cl$\rightarrow$Ar & Cl$\rightarrow$Pr & Cl$\rightarrow$Rw
& Pr$\rightarrow$Ar & Pr$\rightarrow$Cl & Pr$\rightarrow$Rw
& Rw$\rightarrow$Ar & Rw$\rightarrow$Cl & Rw$\rightarrow$Pr
& Avg. \\
\midrule
\multicolumn{15}{l}{SDXL-prior generation variants: prompt-only generation} \\
\cmidrule(lr){1-15}
\rowcolor{elsbg}
ELS+SDXL (class-only)
& Prompt-only & 56.88 & 73.64 & 80.38 & 69.18 & 73.64 & 80.40 & 68.93 & 56.54 & 80.38 & 68.93 & 56.54 & 73.64 & 69.92 \\
\rowcolor{elsbg}
ELS+SDXL (style prompts)
& Prompt-only & 55.23 & 77.09 & 80.26 & 68.11 & 77.09 & 80.26 & 68.11 & 55.23 & 80.45 & 68.11 & 55.23 & 77.04 & 70.18 \\
\rowcolor{elsbg}
ELS+SDXL (target name)
& Prompt-only & 59.70 & 75.51 & 82.26 & 66.67 & 75.51 & \asecond{82.26} & 66.67 & 59.70 & 82.26 & 66.67 & 59.70 & 75.51 & 71.04 \\
\rowcolor{elsbg}
ELS+SDXL (target-style prompts)
& Prompt-only & 60.76 & 79.52 & 81.68 & 70.95 & 79.52 & 81.68 & 70.95 & 60.76 & 81.68 & 70.95 & 60.76 & 79.52 & 73.23 \\
\rowcolor{elsbg}
ELS+SDXL (selected)
& Prompt-only & 61.63 & 79.81 & 82.19 & \asecond{71.98} & 79.73 & 81.82 & 71.69 & 61.58 & 82.07 & 72.76 & 62.15 & 80.42 & 73.99 \\
\midrule
\multicolumn{15}{l}{Diffusion-adapted and downstream references} \\
\cmidrule(lr){1-15}
\rowcolor{elsbg}
ELS~\cite{zhang2023els}
& \textemdash & 57.79 & 77.65 & 81.62 & 66.59 & 76.74 & 76.43 & 62.69 & 56.69 & 82.12 & 75.63 & 62.85 & 85.35 & 71.84 \\
\rowcolor{elsbg}
ELS+Terra~\cite{zhuang2024terra}
& Pairwise & \asecond{64.62} & \asecond{82.33} & \abest{83.60} & 71.19 & \abest{84.25} & 80.31 & \asecond{73.00} & \asecond{63.57} & \asecond{83.81} & \abest{76.20} & \asecond{66.56} & \asecond{85.70} & \asecond{76.26} \\
\rowcolor{elsbg}
ELS+DCDM~\cite{zhang2025dcdm}
& Pairwise & 60.35 & 78.81 & \asecond{82.74} & 69.59 & 80.53 & 79.55 & 65.16 & 58.26 & 83.11 & \asecond{75.81} & 64.18 & 85.55 & 73.64 \\
\midrule
\multicolumn{15}{l}{\textbf{Our method: multi-target generation}} \\
\cmidrule(lr){1-15}
\rowcolor{elsbg}
ELS+MUSE
& Multi-target & \abest{66.51} & \abest{84.09} & 82.62 & \abest{74.14} & \asecond{83.74} & \abest{82.96} & \abest{74.61} & \abest{64.73} & \abest{85.52} & 75.61 & \abest{68.81} & \abest{87.45} & \abest{77.57} \\
\bottomrule
\end{tabular}%
}
\end{table*}

\section{Feature-Space Visualization}
\label{app:tsne_visualization}

\begin{figure*}[t]
    \centering
    \includegraphics[width=0.98\textwidth]{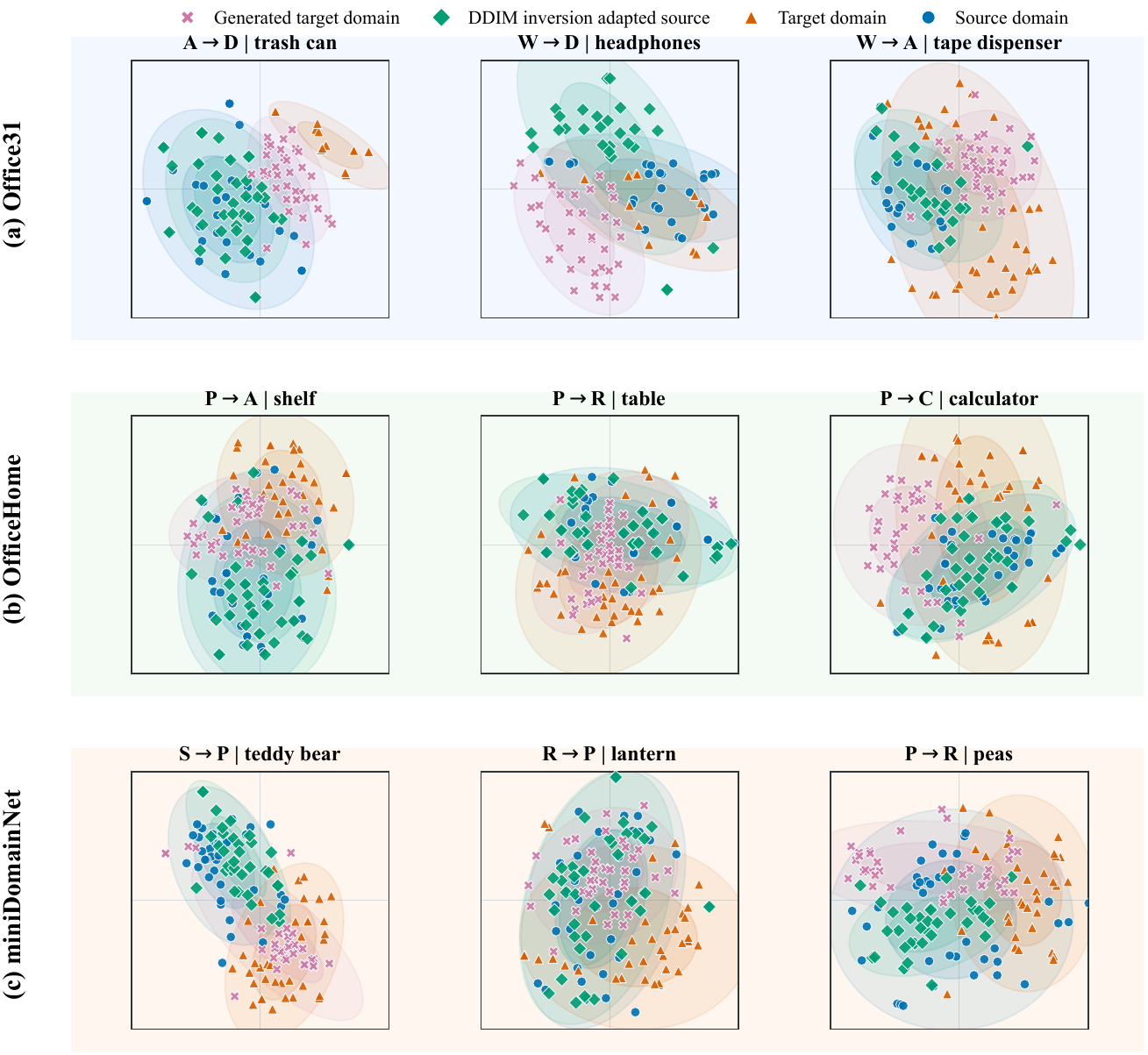}
    \caption{
    t-SNE visualizations of feature distributions for representative categories from Office-31, Office-Home, and miniDomainNet. We compare source-domain samples, target-domain samples, DDIM-inversion adapted source samples, and generated target-domain samples.
    }
    \label{fig:tsne}
\end{figure*}

Figure~\ref{fig:tsne} shows t-SNE~\cite{tsne} visualizations for representative categories from Office-31, Office-Home, and miniDomainNet. We compare four distributions: source-domain samples, target-domain samples, DDIM-inversion adapted source samples, and generated target-domain samples. The generated target-domain samples tend to occupy regions close to the target-domain distribution, while the DDIM-inversion adapted source samples often lie between the source and target domains or overlap with the target clusters. This suggests that the proposed generation process produces target-style samples with improved feature-level alignment, thereby helping reduce the source--target domain gap.

\section{Qualitative Visualization of Generated Bridge Samples}
\label{app:qualitative_visualization}

We provide qualitative visualizations of the generated bridge samples used in our diffusion-based UDA protocol. The visualizations cover all three benchmarks: Office-31, Office-Home, and miniDomainNet. For each benchmark, we show two types of generated data: source-to-target translated samples obtained by DDIM inversion, and class-conditional samples generated from Gaussian noise. The DDIM-inversion visualizations are arranged in paired columns: within each pair, the left image is the original source image and the right image is the corresponding target-style sample generated after inversion and denoising with the target-specific MUSE branch. The pure-noise visualizations show class-conditional target-style samples generated directly from Gaussian noise using the corresponding target-specific branch. These figures are intended as qualitative evidence for the behavior of the generator.

\subsection{Office-31}
\label{app:qualitative_office31}

\begin{figure*}[t]
    \centering
    \includegraphics[width=0.98\textwidth]{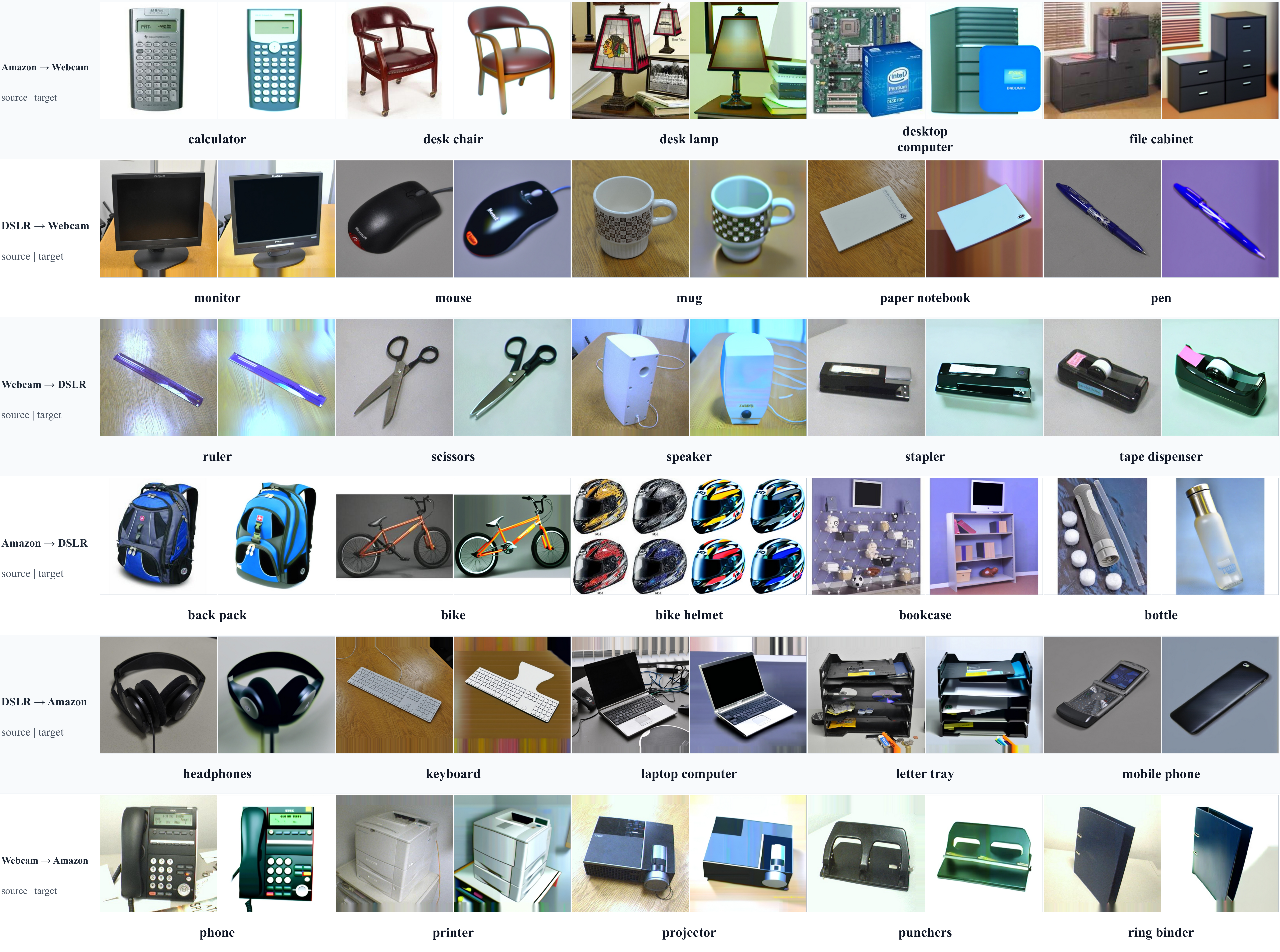}
    \caption{Qualitative visualization of DDIM-inversion-based source-to-target generation on Office-31. Rows correspond to the 6 transfer tasks A$\rightarrow$W, D$\rightarrow$W, W$\rightarrow$D, A$\rightarrow$D, D$\rightarrow$A, and W$\rightarrow$A. Each row shows 5 classes. Within each source--generated pair, the left image is the source-domain input and the right image is the corresponding target-style image generated by the target-specific MUSE branch.}
    \label{fig:office31_ddim_visualization}
\end{figure*}

\begin{figure*}[t]
    \centering
    \includegraphics[width=0.98\textwidth]{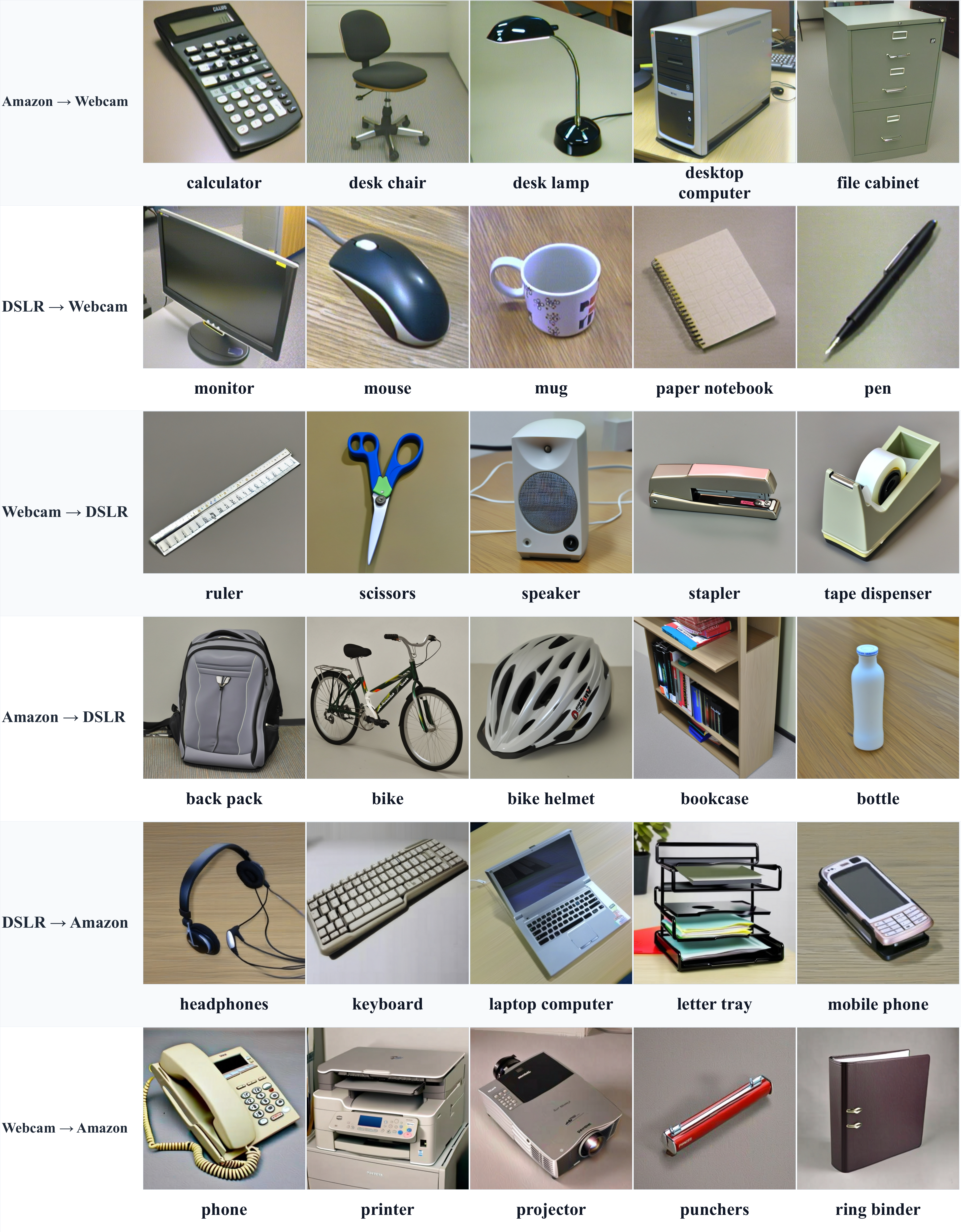}
    \caption{Qualitative visualization of class-conditional samples generated from Gaussian noise on Office-31. Rows correspond to the 6 transfer tasks A$\rightarrow$W, D$\rightarrow$W, W$\rightarrow$D, A$\rightarrow$D, D$\rightarrow$A, and W$\rightarrow$A. Each row shows generated target-style samples from five different classes.}
    \label{fig:office31_noise_visualization}
\end{figure*}

Figure~\ref{fig:office31_ddim_visualization} visualizes DDIM-inversion-based source-to-target translation on Office-31. The rows correspond to the six ordered transfer tasks, following the order used in Table~\ref{tab:office31_main}: A$\rightarrow$W, D$\rightarrow$W, W$\rightarrow$D, A$\rightarrow$D, D$\rightarrow$A, and W$\rightarrow$A. For each transfer task, five different classes are shown. Each displayed example is organized as a source--generated pair, where the left image is the source-domain input and the right image is the generated target-style image. Figure~\ref{fig:office31_noise_visualization} shows samples generated from Gaussian noise on Office-31. The same six transfer tasks are displayed from top to bottom, and each row contains target-style class-conditional samples from five different classes.

\subsection{Office-Home}
\label{app:qualitative_officehome}

\begin{figure*}[t]
    \centering
    \includegraphics[width=0.85\textwidth]{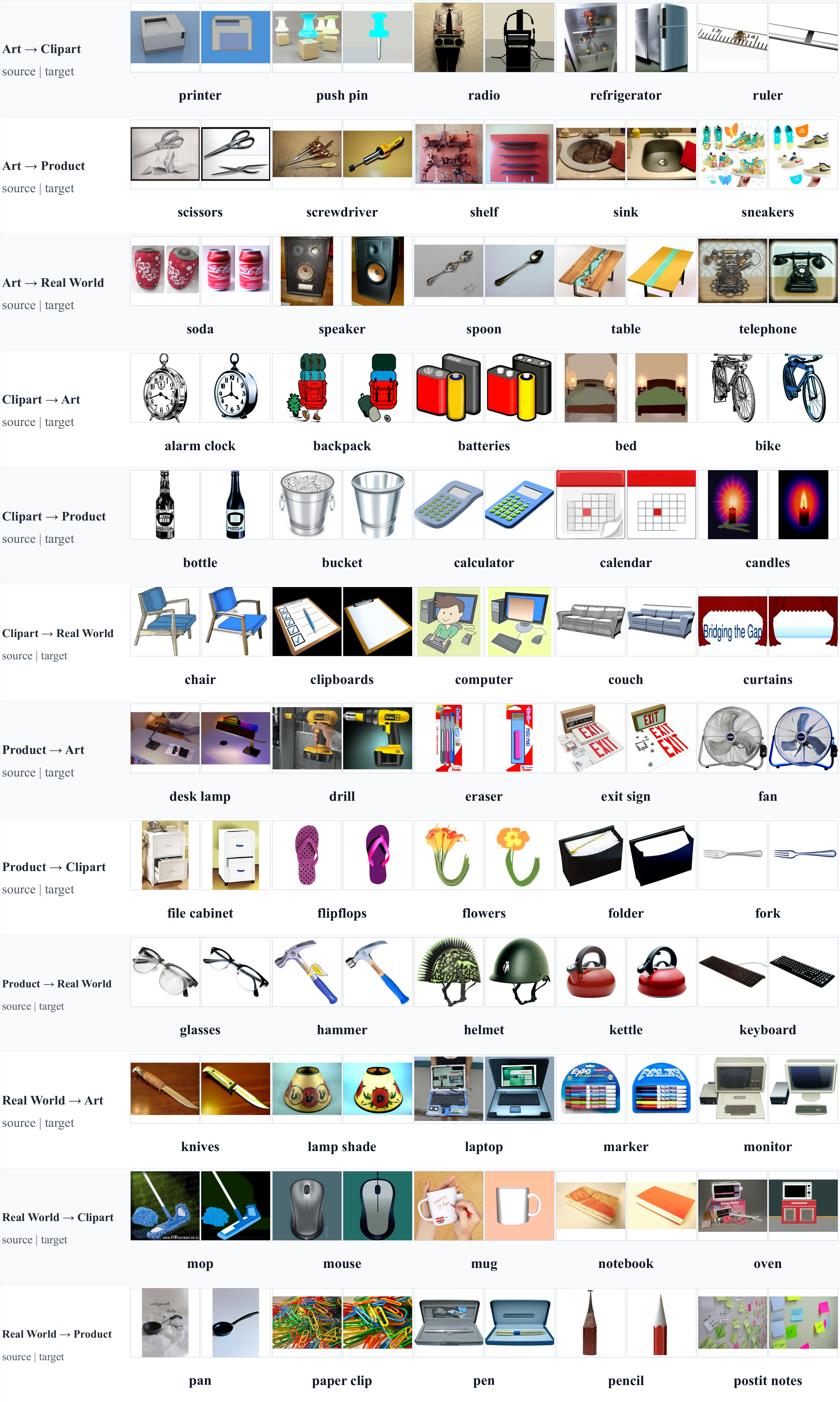}
    \caption{Qualitative visualization of DDIM-inversion-based source-to-target generation on Office-Home. Rows correspond to the 12 transfer tasks Ar$\rightarrow$Cl, Ar$\rightarrow$Pr, Ar$\rightarrow$Rw, Cl$\rightarrow$Ar, Cl$\rightarrow$Pr, Cl$\rightarrow$Rw, Pr$\rightarrow$Ar, Pr$\rightarrow$Cl, Pr$\rightarrow$Rw, Rw$\rightarrow$Ar, Rw$\rightarrow$Cl, and Rw$\rightarrow$Pr. Each row shows 5 classes. Within each pair, the left image is the source-domain input and the right image is the generated target-style image.}
    \label{fig:officehome_ddim_visualization}
\end{figure*}

\begin{figure*}[t]
    \centering
    \includegraphics[width=0.55\textwidth]{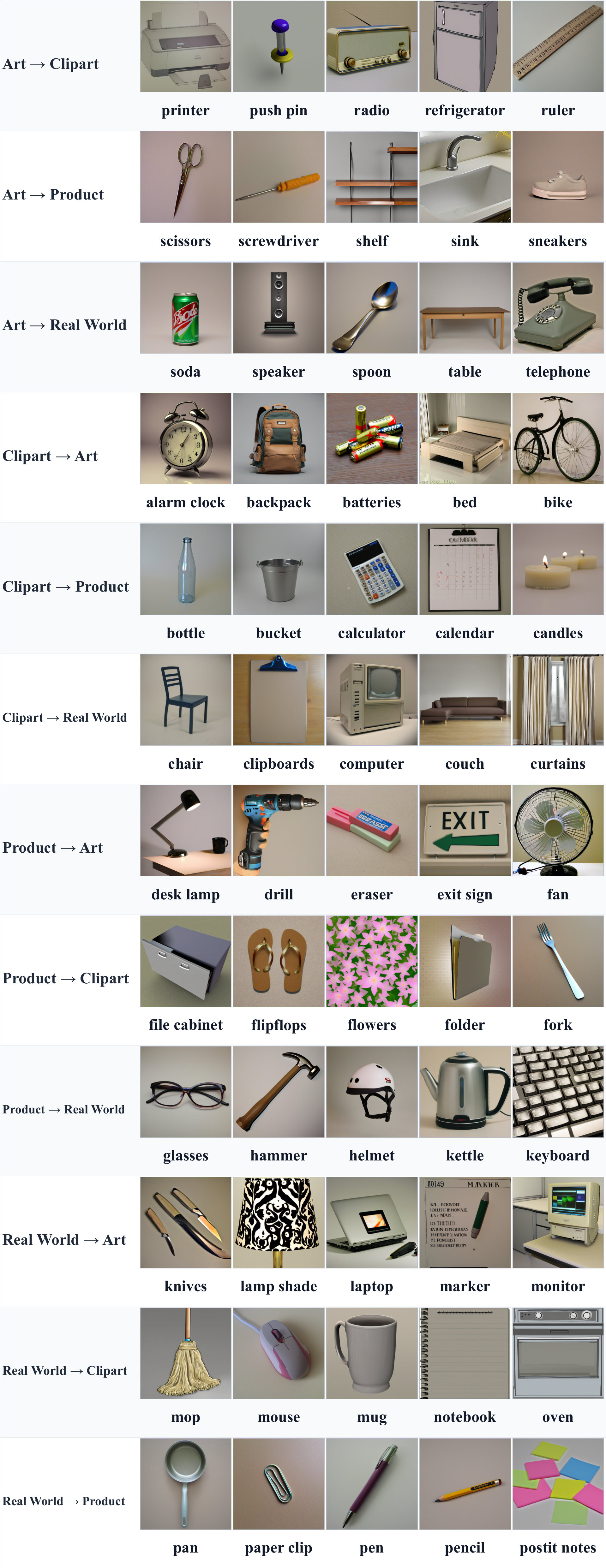}
    \caption{Qualitative visualization of class-conditional samples generated from Gaussian noise on Office-Home. Rows correspond to the 12 transfer tasks Ar$\rightarrow$Cl, Ar$\rightarrow$Pr, Ar$\rightarrow$Rw, Cl$\rightarrow$Ar, Cl$\rightarrow$Pr, Cl$\rightarrow$Rw, Pr$\rightarrow$Ar, Pr$\rightarrow$Cl, Pr$\rightarrow$Rw, Rw$\rightarrow$Ar, Rw$\rightarrow$Cl, and Rw$\rightarrow$Pr. Each row shows generated target-style samples from 5 different classes.}
    \label{fig:officehome_noise_visualization}
\end{figure*}

Figure~\ref{fig:officehome_ddim_visualization} visualizes DDIM-inversion-based source-to-target translation on Office-Home. The rows correspond to the twelve ordered transfer tasks in Table~\ref{tab:officehome_main}: Ar$\rightarrow$Cl, Ar$\rightarrow$Pr, Ar$\rightarrow$Rw, Cl$\rightarrow$Ar, Cl$\rightarrow$Pr, Cl$\rightarrow$Rw, Pr$\rightarrow$Ar, Pr$\rightarrow$Cl, Pr$\rightarrow$Rw, Rw$\rightarrow$Ar, Rw$\rightarrow$Cl, and Rw$\rightarrow$Pr. For each transfer task, five different classes are shown. The paired layout follows the same convention as above: the left image in each pair is the source image, and the right image is the generated target-style image. Figure~\ref{fig:officehome_noise_visualization} shows the corresponding pure-noise generation results. Each row corresponds to one transfer task and contains class-conditional target-style samples from five different classes.

\subsection{miniDomainNet}
\label{app:qualitative_minidomainnet}

\begin{figure*}[t]
    \centering
    \includegraphics[width=0.98\textwidth]{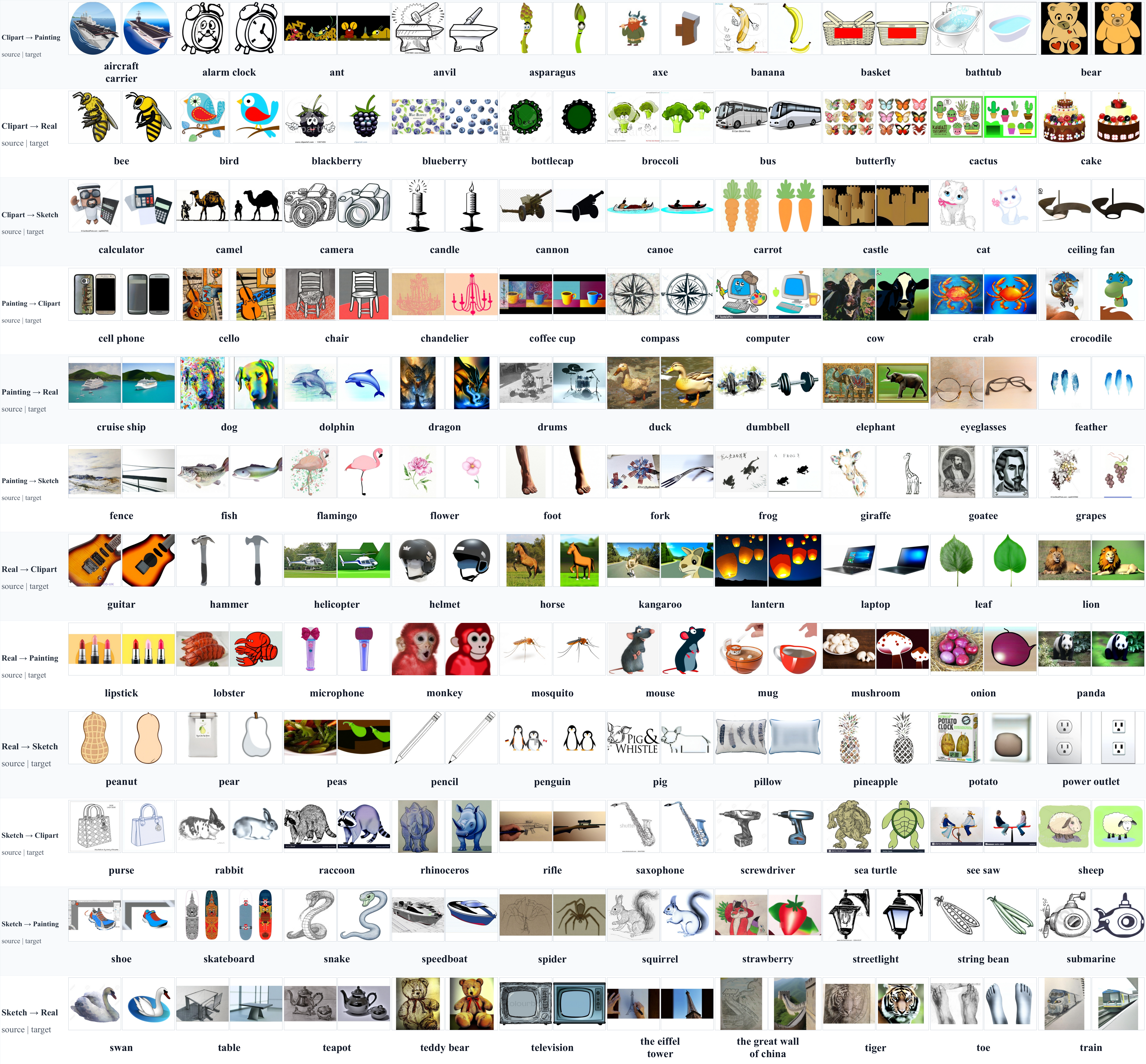}
    \caption{Qualitative visualization of DDIM-inversion-based source-to-target generation on miniDomainNet. Rows correspond to the 12 transfer tasks C$\rightarrow$P, C$\rightarrow$R, C$\rightarrow$S, P$\rightarrow$C, P$\rightarrow$R, P$\rightarrow$S, R$\rightarrow$C, R$\rightarrow$P, R$\rightarrow$S, S$\rightarrow$C, S$\rightarrow$P, and S$\rightarrow$R. Each row shows 10 classes. Within each pair, the left image is the source-domain input and the right image is the corresponding target-style image generated by the target-specific MUSE branch.}
    \label{fig:minidomainnet_ddim_visualization}
\end{figure*}

\begin{figure*}[t]
    \centering
    \includegraphics[width=0.95\textwidth]{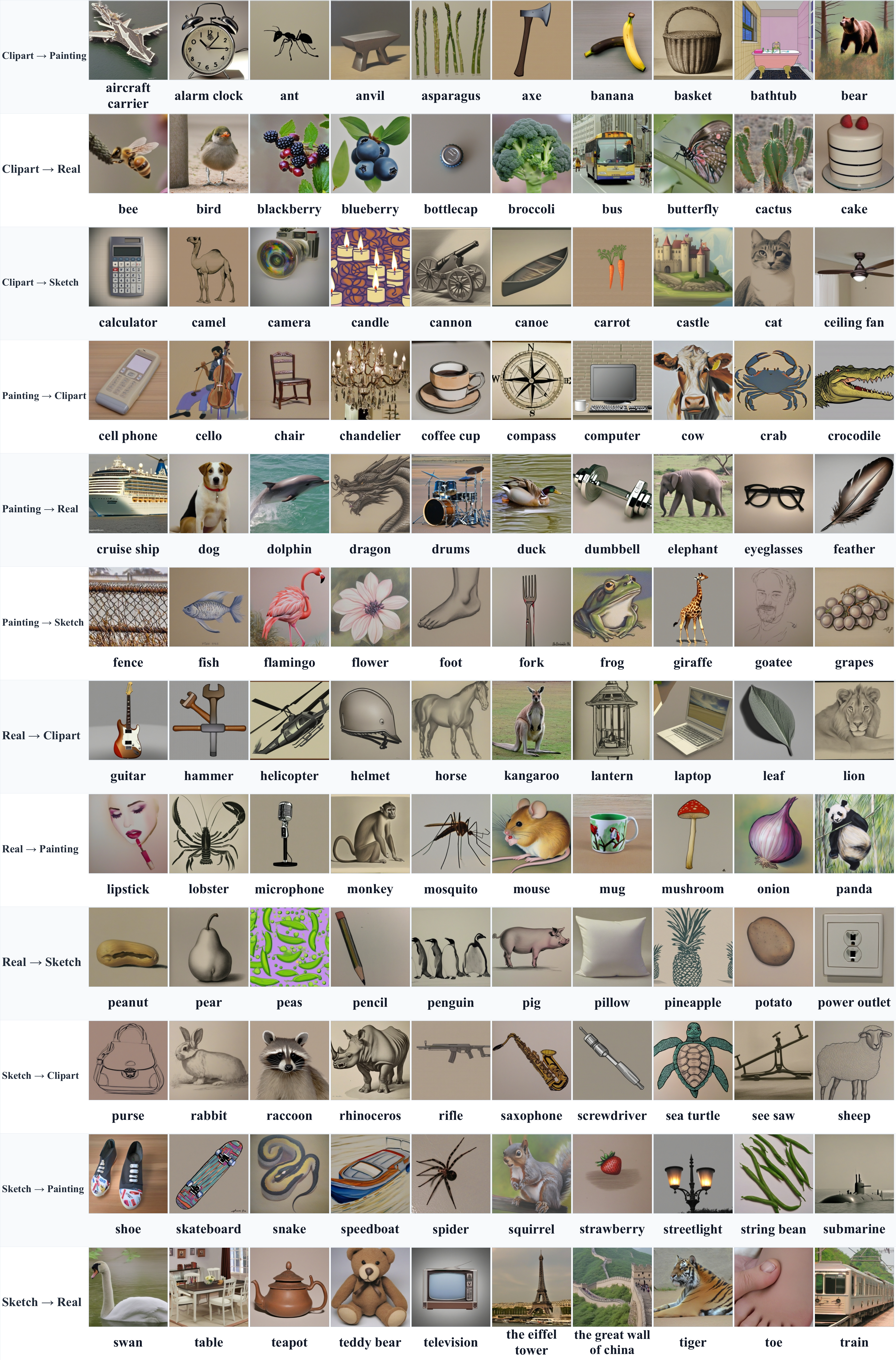}
    \caption{Qualitative visualization of class-conditional samples generated from Gaussian noise on miniDomainNet. Rows correspond to the 12 transfer tasks C$\rightarrow$P, C$\rightarrow$R, C$\rightarrow$S, P$\rightarrow$C, P$\rightarrow$R, P$\rightarrow$S, R$\rightarrow$C, R$\rightarrow$P, R$\rightarrow$S, S$\rightarrow$C, S$\rightarrow$P, and S$\rightarrow$R. Each row shows generated target-style samples from 10 different classes.}
    \label{fig:minidomainnet_noise_visualization}
\end{figure*}

Figure~\ref{fig:minidomainnet_ddim_visualization} visualizes DDIM-inversion-based source-to-target translation on miniDomainNet. The rows correspond to the twelve ordered transfer tasks in Table~\ref{tab:minidomainnet_main}: C$\rightarrow$P, C$\rightarrow$R, C$\rightarrow$S, P$\rightarrow$C, P$\rightarrow$R, P$\rightarrow$S, R$\rightarrow$C, R$\rightarrow$P, R$\rightarrow$S, S$\rightarrow$C, S$\rightarrow$P, and S$\rightarrow$R. For each transfer task, ten different classes are shown. As in the other DDIM-inversion figures, each example is presented as a source--generated pair, with the source image on the left and the generated target-style image on the right. Figure~\ref{fig:minidomainnet_noise_visualization} shows class-conditional pure-noise generation on miniDomainNet. Each row corresponds to one transfer task and contains generated target-style samples from ten different classes.

\section{Limitations}
\label{sec:limitations}

Despite its empirical effectiveness, MUSE has several limitations. First, the method is developed under the closed-set UDA assumption, where the source and all target domains share the same label space. This assumption is standard in many UDA benchmarks, but it may not hold in practical applications where target domains contain unknown classes, missing classes, or class-prior shifts. In such cases, class-conditional generation using source labels may introduce biased or incorrectly labeled synthetic supervision.

Second, MUSE reduces the cost of repeated diffusion fine-tuning but does not eliminate the computational burden of diffusion-based data generation. The method still relies on a large SDXL backbone, high-resolution image preprocessing, and iterative sampling through DDIM inversion or class-conditional generation. Therefore, compared with purely discriminative UDA methods, MUSE remains more expensive in terms of GPU memory, training time, and sampling time, which may limit its use in resource-constrained environments.

Third, the branch-decoupled design of MUSE provides an operational separation between source-supervised semantic adaptation and target-specific appearance adaptation, but it does not assume that image semantics and style are perfectly independent. This separation can become less precise when the domain shift itself changes label-relevant visual structure, such as under large viewpoint changes, occlusion, extreme abstraction, or strongly class-dependent target styles. In such cases, the target-style branch may also affect class-discriminative visual details. In addition, when the pretrained diffusion prior or the class prompt provides weak support for a particular category or target domain, individual generated samples may exhibit low-level artifacts, incomplete target-style transfer, or occasional semantic drift.

In the evaluated benchmarks, these generation imperfections do not undermine the overall effectiveness of the generated bridge sets: MUSE consistently improves average downstream UDA accuracy over the corresponding discriminative baselines and repeated diffusion-based adaptation. Several design choices also reduce sensitivity to individual generation errors. DDIM inversion preserves source-instance structure, class-conditioned generation reinforces category information, target-domain losses are prevented from directly updating the shared semantic branch, and the final bridge set combines inversion-based and class-conditional samples to provide complementary synthetic supervision. Nevertheless, generation quality remains dependent on the pretrained diffusion prior and the severity of the source--target shift, and handling more extreme structural domain shifts is an important direction for future work.

\section{Broader Impacts}
MUSE studies diffusion-based synthetic-data generation for unsupervised domain adaptation. A potential positive impact is improving the practicality of diffusion-based UDA when one labeled source domain must support multiple unlabeled target domains, by reducing repeated source-guided diffusion fine-tuning and storage relative to independent per-target adaptation. This may benefit applications where labeled target-domain data are difficult to obtain and multiple visual domains need to be handled. However, MUSE still relies on a large diffusion backbone and iterative generation, so it should not be interpreted as eliminating the computational cost of diffusion-based adaptation. Because the method builds on image generation models, generated data could also be misused if applied outside controlled benchmark settings, for example to synthesize misleading visual content or to amplify dataset biases. In this work, we evaluate only object-centric academic benchmarks and do not release a general-purpose image generator. Future deployments should consider dataset bias, provenance of generated samples, labeling errors, computational cost, and safeguards against misuse.

\section{Existing Assets and Licenses}
We use existing datasets, pretrained models, and baseline implementations only for academic research purposes. We cite the original works for Office-31, Office-Home, miniDomainNet, SDXL, and all compared UDA methods. For the existing datasets, model checkpoints, and codebases used in the experiments, we checked the corresponding official sources, licenses, and terms of use, and used them in accordance with their stated conditions. We do not scrape new data or repackage existing datasets beyond the data preparation needed for the reported experiments. The released supplemental code and data are anonymized for submission and include documentation of the assets needed to reproduce the experiments.

\clearpage

\newpage
\section*{NeurIPS Paper Checklist}

\begin{enumerate}

\item {\bf Claims}
    \item[] Question: Do the main claims made in the abstract and introduction accurately reflect the paper's contributions and scope?
    \item[] Answer: \answerYes{}
    \item[] Justification: The abstract and introduction state the paper's scope as closed-set multi-target data generation for diffusion-based UDA, and the main claims are supported by the method formulation in Sec.~\ref{sec:method}, the experimental results in Sec.~\ref{sec:experiments}, and the efficiency analysis. The assumptions and limitations of the setting are further discussed in Sec.~\ref{sec:limitations}.

\item {\bf Limitations}
    \item[] Question: Does the paper discuss the limitations of the work performed by the authors?
    \item[] Answer: \answerYes{}
    \item[] Justification: The paper includes a dedicated Limitations section in Sec.~\ref{sec:limitations}, discussing the closed-set assumption, the computational cost of SDXL-based diffusion adaptation and sampling, and possible generation artifacts or semantic drift.

\item {\bf Theory assumptions and proofs}
    \item[] Question: For each theoretical result, does the paper provide the full set of assumptions and a complete (and correct) proof?
    \item[] Answer: \answerYes{}
    \item[] Justification: The theoretical results in the paper are limited to the adapter-parameter scaling analysis. Appendix~\ref{app:muse_scaling} states the assumptions, defines the parameterization, and provides complete proofs for the proposition and corollary.

\item {\bf Experimental result reproducibility}
    \item[] Question: Does the paper fully disclose all the information needed to reproduce the main experimental results of the paper to the extent that it affects the main claims and/or conclusions of the paper (regardless of whether the code and data are provided or not)?
    \item[] Answer: \answerYes{}
    \item[] Justification: The paper describes the datasets, metrics, compared methods, generation protocol, downstream UDA protocol, and implementation details in Sec.~\ref{sec:experiments} and Appendix~\ref{app:implementation_details}. These include preprocessing, adapter placement and ranks, training losses, optimizer settings, sampling strategy, random seed, and generation settings needed to reproduce the main results.

\item {\bf Open access to data and code}
    \item[] Question: Does the paper provide open access to the data and code, with sufficient instructions to faithfully reproduce the main experimental results, as described in supplemental material?
    \item[] Answer: \answerYes{}
    \item[] Justification: The supplemental material provides anonymized code and data with instructions for reproducing the main experimental results, including data preparation, diffusion-adapter training, bridge-set generation, and downstream UDA evaluation. The paper also provides detailed reproduction information in Sec.~\ref{sec:experiments} and Appendix~\ref{app:implementation_details}.

\item {\bf Experimental setting/details}
    \item[] Question: Does the paper specify all the training and test details (e.g., data splits, hyperparameters, how they were chosen, type of optimizer) necessary to understand the results?
    \item[] Answer: \answerYes{}
    \item[] Justification: Sec.~\ref{sec:experiments} specifies the benchmarks, metrics, compared methods, and evaluation protocol, while Appendix~\ref{app:implementation_details} provides the training and sampling details, including preprocessing, adapter ranks, optimizer, learning rates, batch size, training steps, random seed, prompts, and target-domain sampling.

\item {\bf Experiment statistical significance}
    \item[] Question: Does the paper report error bars suitably and correctly defined or other appropriate information about the statistical significance of the experiments?
    \item[] Answer: \answerYes{}
    \item[] Justification: The main experimental results are accompanied by standard deviations over three independent runs, reported in Appendix~\ref{app:standard_deviations}. The appendix states that the reported variability corresponds to run-to-run variation under the same evaluation protocol.

\item {\bf Experiments compute resources}
    \item[] Question: For each experiment, does the paper provide sufficient information on the computer resources (type of compute workers, memory, time of execution) needed to reproduce the experiments?
    \item[] Answer: \answerYes{}
    \item[] Justification: The paper reports the main compute resources in Sec.~\ref{sec:experiments} and Appendix~\ref{app:implementation_details}, including the use of one NVIDIA A100 80GB GPU, 16 CPU cores, diffusion fine-tuning time, and per-image bridge-generation latency. Table~\ref{tab:finetune_time} reports the fine-tuning time for the compute-heavy diffusion adaptation stage.
    
\item {\bf Code of ethics}
    \item[] Question: Does the research conducted in the paper conform, in every respect, with the NeurIPS Code of Ethics \url{https://neurips.cc/public/EthicsGuidelines}?
    \item[] Answer: \answerYes{}
    \item[] Justification: The research uses standard public UDA benchmarks and pretrained diffusion backbones, and does not involve human subjects, private personal data, or deployment in high-stakes decision-making settings. We have reviewed the NeurIPS Code of Ethics and believe the work conforms to it.

\item {\bf Broader impacts}
    \item[] Question: Does the paper discuss both potential positive societal impacts and negative societal impacts of the work performed?
    \item[] Answer: \answerYes{}
    \item[] Justification: The paper includes a Broader Impacts section that discusses the potential positive impact of reducing the cost of multi-target adaptation and the potential negative impacts associated with synthetic image generation, dataset bias, and misuse outside controlled benchmark settings.
    
\item {\bf Safeguards}
    \item[] Question: Does the paper describe safeguards that have been put in place for responsible release of data or models that have a high risk for misuse (e.g., pre-trained language models, image generators, or scraped datasets)?
    \item[] Answer: \answerNA{}
    \item[] Justification: The paper does not release a high-risk pretrained model, a general-purpose image generator, or a scraped dataset. The released supplemental assets are intended to reproduce controlled object-centric UDA benchmark experiments rather than to provide a general-purpose generation system.

\item {\bf Licenses for existing assets}
    \item[] Question: Are the creators or original owners of assets (e.g., code, data, models), used in the paper, properly credited and are the license and terms of use explicitly mentioned and properly respected?
    \item[] Answer: \answerYes{}
    \item[] Justification: We cite the original creators of all datasets, pretrained models, and baseline methods used in the paper. The Existing Assets and Licenses section states that we checked the corresponding official sources, licenses, and terms of use and used the assets in accordance with their stated conditions.

\item {\bf New assets}
    \item[] Question: Are new assets introduced in the paper well documented and is the documentation provided alongside the assets?
    \item[] Answer: \answerYes{}
    \item[] Justification: The supplemental material provides the new code and data associated with the paper, together with documentation for data preparation, training, bridge-set generation, downstream evaluation, and reproduction of the main experimental results. The released assets are anonymized for submission and documented alongside the supplemental material.

\item {\bf Crowdsourcing and research with human subjects}
    \item[] Question: For crowdsourcing experiments and research with human subjects, does the paper include the full text of instructions given to participants and screenshots, if applicable, as well as details about compensation (if any)? 
    \item[] Answer: \answerNA{}
    \item[] Justification: The paper does not involve crowdsourcing, user studies, human evaluation, or research with human subjects.

\item {\bf Institutional review board (IRB) approvals or equivalent for research with human subjects}
    \item[] Question: Does the paper describe potential risks incurred by study participants, whether such risks were disclosed to the subjects, and whether Institutional Review Board (IRB) approvals (or an equivalent approval/review based on the requirements of your country or institution) were obtained?
    \item[] Answer: \answerNA{}
    \item[] Justification: The paper does not involve crowdsourcing or research with human subjects, so IRB approval or equivalent review is not applicable.

\item {\bf Declaration of LLM usage}
    \item[] Question: Does the paper describe the usage of LLMs if it is an important, original, or non-standard component of the core methods in this research? Note that if the LLM is used only for writing, editing, or formatting purposes and does \emph{not} impact the core methodology, scientific rigor, or originality of the research, declaration is not required.
    \item[] Answer: \answerNA{}
    \item[] Justification: The core method development of MUSE does not involve LLMs as an important, original, or non-standard component. GPT-4 is used only for generating prompts in an auxiliary prompt-only SDXL ablation, and this usage is described in Appendix~\ref{app:ablation_sdxl_prior}.

\end{enumerate}

\end{document}